\documentclass[preprint,5p,times,twocolumn]{elsarticle}
\usepackage{times}  
\usepackage{helvet}  
\usepackage{courier}  
\usepackage[hyphens]{url}  
\usepackage{graphicx} 
\usepackage{amsmath}
\usepackage{bm}
\usepackage{multicol}
\usepackage{rotating}
\usepackage{multirow}
\usepackage{subfigure}
\usepackage{amsthm,amssymb,amsfonts}
\usepackage{booktabs}
\usepackage[inkscapelatex=false]{svg}
\usepackage{float}
\usepackage{algorithm}
\usepackage{algorithmic}
\usepackage{newfloat}
\usepackage{listings}
\usepackage{newfloat}
\usepackage{listings}

\usepackage{amssymb}

\journal{Neural Networks}

\begin{document}

\begin{frontmatter}



\title{FCDNet: Frequency-Guided Complementary Dependency Modeling for Multivariate Time-Series Forecasting}


\author[PKU]{Weijun Chen}

\author[PKU]{Heyuan Wang\corref{cor1}}

\author[PKU]{Ye Tian}

\author[PKU]{Shijie Guan}

\author[PKU]{Ning Liu}

\cortext[cor1]{Corresponding author}

\affiliation[PKU]{organization={School of Computer Science, Peking University},
            state={Beijing},
            country={China}}

\begin{abstract}
Multivariate time-series (MTS) forecasting is a challenging task in many real-world non-stationary dynamic scenarios. In addition to intra-series temporal signals,  the inter-series dependency also plays a crucial role in shaping future trends. How to enable the model's awareness of dependency information has raised substantial research attention. Previous approaches have either presupposed dependency constraints based on domain knowledge or imposed them using real-time feature similarity. However, MTS data often exhibit both enduring long-term static relationships and transient short-term interactions, which mutually influence their evolving states. It is necessary to recognize and incorporate the complementary dependencies for more accurate MTS prediction. The frequency information in time series reflects the evolutionary rules behind complex temporal dynamics, and different frequency components can be used to well construct long-term and short-term interactive dependency structures between variables. To this end, we propose FCDNet, a concise yet effective framework for multivariate time-series forecasting. Specifically, FCDNet overcomes the above limitations by applying two light-weight dependency constructors to help extract long- and short-term dependency information adaptively from multi-level frequency patterns. With the growth of input variables, the number of trainable parameters in FCDNet only increases linearly, which is conducive to the model's scalability and avoids over-fitting. Additionally, adopting a frequency-based perspective can effectively mitigate the influence of noise within MTS data, which helps capture more genuine dependencies. The experimental results on six real-world datasets from multiple fields show that FCDNet significantly exceeds strong baselines, with an average improvement of 6.82$\%$ on MAE, 4.98$\%$ on RMSE, and 4.91$\%$ on MAPE. In addition, the ability of FCDNet to jointly learn high-quality static and dynamic graph structures is demonstrated empirically. Our source codes are publicly available at https://github.com/onceCWJ/FCDNet.
\end{abstract}


\end{frontmatter}


\section{Introduction}
Multivariate time-series forecasting is an important task in the data mining field and has been widely explored in many dynamic scenarios such as financial investment, transportation, and disease prevention~\cite{GMAN,StemGNN,fiancial}. For example, profitable investment strategies can be formulated by predicting future stock price movements, and the government can make early warnings to save lives by analyzing the epidemic trend. Traditional algorithms such as ARIMA~\cite{ARIMA} and state space models~\cite{SSM} are not suitable for many situations due to their strict requirement on the stationarity of a time-series. With the development of deep learning, some research devotes to designing advanced neural networks to extract nonlinear and implicit temporal patterns shared by MTS, such as FC-LSTM~\cite{FC-LSTM} and LSTNet~\cite{LSTNet}. However, they treat each time-series in isolation while ignoring an important property of MTS, that variables interact and co-evolve with each other. For example, stocks of listed companies from the same sector tend to exhibit synchronous trends~\cite{stock_relation}, and records of traffic sensors in the same road network have strong correlations~\cite{GMAN}. 

To perceive the relational dependencies between variables, recent studies~\cite{DCRNN,stock_relation} introduce graph neural networks (GNNs)~\cite{GCN,GAT} into MTS forecasting. However, these GNNs require pre-defined adjacency graphs, which are difficult to obtain in the absence of domain expert knowledge. Meanwhile, static graphs cannot reflect the short-term dynamic dependency in practical complex systems. In this regard, adaptive graph structure learning in a data-driven manner raises great attention. In most current models such as Graph WaveNet (GWN)~\cite{GWN}, Adaptive Graph Convolutional Recurrent Network (AGCRN)~\cite{AGCRN}, and MTGNN~\cite{MTGNN}, the graph structure is randomly initialized and tuned end-to-end during model training. Once the training of the model finishes, the dependency matrix is fixed. Therefore, these models still use a static modeling process for dynamic input MTS. However, predicting with static graphs causes significant bias because the correlations between variables are time-varying in the real world. Another paradigm to tackle such a problem is dynamic graph learning~\cite{stock_attention,DSTAGNN,Aware}. However, despite achieving appreciable improvements, they often depend on highly complex architectures to capture subtle trends and interdependent information contained in MTS, which incurs significant time costs and quadratic memory consumption, thus limiting the model's scalability.

Existing methods have made significant strides in modeling dependencies within MTS data. However, they often overlook the fact that MTS possess both long-term stable dependencies and short-term, immediate interactions. In various domains, such as the securities market or traffic scenarios, these time series exhibit long-term patterns, like sector rotation in stocks or constant distance relationships between road points. Yet, they also experience short-term variations caused by events like emergencies in the stock market or temporary changes in traffic flow during holidays and peak hours. To enhance the accuracy of predictions in MTS data, models must strike a balance between two complementary aspects: accurately learning long-term dependencies and sensibly capturing short-term changes in relationships. This balance allows the model to effectively capture both stable, long-term patterns and dynamic, short-term fluctuations, leading to more precise and robust predictions.

Besides, real-world MTS is usually collected from hybrid dynamic systems that blend various signal frequencies and have a low signal-to-noise ratio. Therefore, time-frequency mining is a fundamentally effective solution to help reduce the impact of noise in MTS and discover more robust volatility patterns. For example, in stock price prediction, SFM~\cite{stock1} proposes a state frequency memory recurrent network to capture the multi-frequency trading patterns from past market data to make regression over time. In long-term forecasting~\cite{Autoformer}, time-frequency analysis is applied for robust multiscale time-series feature extraction. FEDFormer~\cite{FedFormer} develops a frequency-enhanced transformer to reduce input and output distribution differences and enhance robustness to noise. In terms of model lightweight, FiLM~\cite{Film} proposes a frequency-improved Legendre memory model for preserving historical information in neural networks while avoiding overfitting to noise presented in history. In general, these methods illustrate the role of frequency in reflecting the periodicity and trend of time series in the temporal domain, but how to combine the frequency method with improving the modeling of dependency structure among multivariate is still a challenge.

Time series data typically exhibit multiple fluctuation components, encompassing high-frequency elements that undergo rapid fluctuations within short time spans, and low-frequency components that change gradually over extended periods. Decomposing time series enables us to discern these fluctuations at different scales, offering a deeper insight into the underlying data structure. The low-frequency component often signifies the long-term trend embedded within the time series. By extracting these low-frequency components, we gain a clearer perspective on trend variations in the data, aiding in the prediction of future long-term trends. Utilizing these low-frequency components facilitates the understanding of stable, long-term correlations between variables. On the other hand, high-frequency components typically encapsulate random fluctuations and noise present in time series data, along with more intricate details. Leveraging the stable correlations derived from low-frequency components allows for adaptive noise reduction and comprehensive exploration of detailed information within high-frequency components. This approach ensures effective and reliable complementary relational analysis.

To this end, in this paper, we propose a concise yet practical dependency modeling framework for MTS forecasting. Unlike previous methods, our model focuses on capturing multi-level frequency temporal information to guide the automatic construction of dependency graphs. In our paper, we show the powerful ability to combine time-frequency mining with complementary dependency modeling for MTS forecasting. The main contributions of our work are as follows:

\begin{itemize}
\item To combine time-frequency mining with structure modeling, we propose a model called FCDNet, which follows a novel route with the aim of learning the complementary effects of long-term dependencies and short-term interactions from multi-level frequency information.

\item Our model develops the \emph{\textbf{L}ong-\textbf{T}erm \textbf{F}requency \textbf{E}xtractor} (LTFE) with wavelet transform and the \emph{\textbf{S}hort-\textbf{T}erm \textbf{F}requency \textbf{E}xtractor} (STFE) with Fast Fourier Transform to emphasize the span of MTS when extracting multi-level frequency information. LTFE learns stable static correlations from long-term historical MTS, and STFE learns dynamic evolving correlations from short-term input MTS. LTFE and STFE are complementary roles that help capture a more comprehensive variable interaction pattern.

\item We conduct extensive experiments on six real-world multivariate time-series datasets, including Ashare, Solar-Energy, PEMS03, PEMS07, PEMS04, and PEMS08. Comparison results with many strong baselines demonstrate the effectiveness and efficiency of our model.
\end{itemize}

\section{Related Work}
The joint modeling of inter-series correlations and intra-series dynamics are critical for MTS forecasting, which means that a variable's future information depends not only on its historical information but also on the historical information of other variables. Spatial-temporal graph neural networks have successfully utilized this characteristic and achieved success in MTS forecasting. The input of spatial-temporal graph networks are usually an MTS and an additionally given adjacency matrix. They aim to predict future values or labels of multivariate time-series. DCRNN~\cite{DCRNN}, STGCN~\cite{STGCN}, and GWN~\cite{GWN} are three representative works for early research on spatial-temporal modeling. After these three works, a series of spatial-temporal models were proposed and achieved success at that time~\cite{GMAN,ASTGCN,MRA-BGCN}. Despite progresses made in the above models, they heavily rely on heuristic rules or a prior-fixed graph structure, which can only store limited correlation information due to the lack of evolving time-series properties, meanwhile inevitably preventing the model from flexible applicability. 

To get rid of the dependence on prior knowledge, recent works tend to automatically model correlations in a data-driven way. For example, MTGNN~\cite{MTGNN} extends GWN by proposing a more delicate structure learning module. AGCRN~\cite{AGCRN} proposes two adaptive modules for enhancing graph convolutional networks to infer dynamic spatial dependencies among different traffic series. Another line of research deems that it is necessary to imitate the dynamics of inter-series relationships more intensively. They propose a variety of architectures~\cite{DMSTGNN,DSTAGNN,DGRNN,Aware,StemGNN,evolution_graph} to formulate a fine topology of dynamic graphs at each time step. DSTAGNN~\cite{DSTAGNN} propose a dynamic spatial-temporal aware graph based on
a data-driven strategy to replace the pre-defined static graph. ST-WA~\cite{Aware} aims at turning spatio-temporal agnostic models
to spatio-temporal aware models by generating location-specific and time-varying model parameters to turn spatio-temporal agnostic models to spatio-temporal aware
models. Though bringing improvement, these methods are not cost-effective since the design of model architectures is extremely sophisticated. This will cause a bottleneck when applied to scalability scenarios.

However, the above methods focus too much on the dynamic interaction between variables and ignore the complementary role of stable correlations between variables. It is non-trivial to simultaneously learn high-quality long-range stable correlations and accurate short-term interactions along with their complementary role. In addition, the above models have not focused on the particularity of structure recognition and modeling in MTS, which differs from general graph modeling problems. ST-Norm~\cite{STNorm} shares a new insight that spatial-temporal indistinguishability is the key difficulty of MTS forecasting and emphasizes that the inherent frequency information is vital to solving the problem. The existing frequency-based methods, such as FEDformer~\cite{FedFormer} and FiLM~\cite{Film}, are all aimed at the temporal evolution of time series. Whereas in our paper, we combine time-frequency information with dependency modeling for MTS forecasting. 

\begin{figure*}
    \centering
    \vspace{-0.3cm}
    \includegraphics[width=1.01\textwidth]{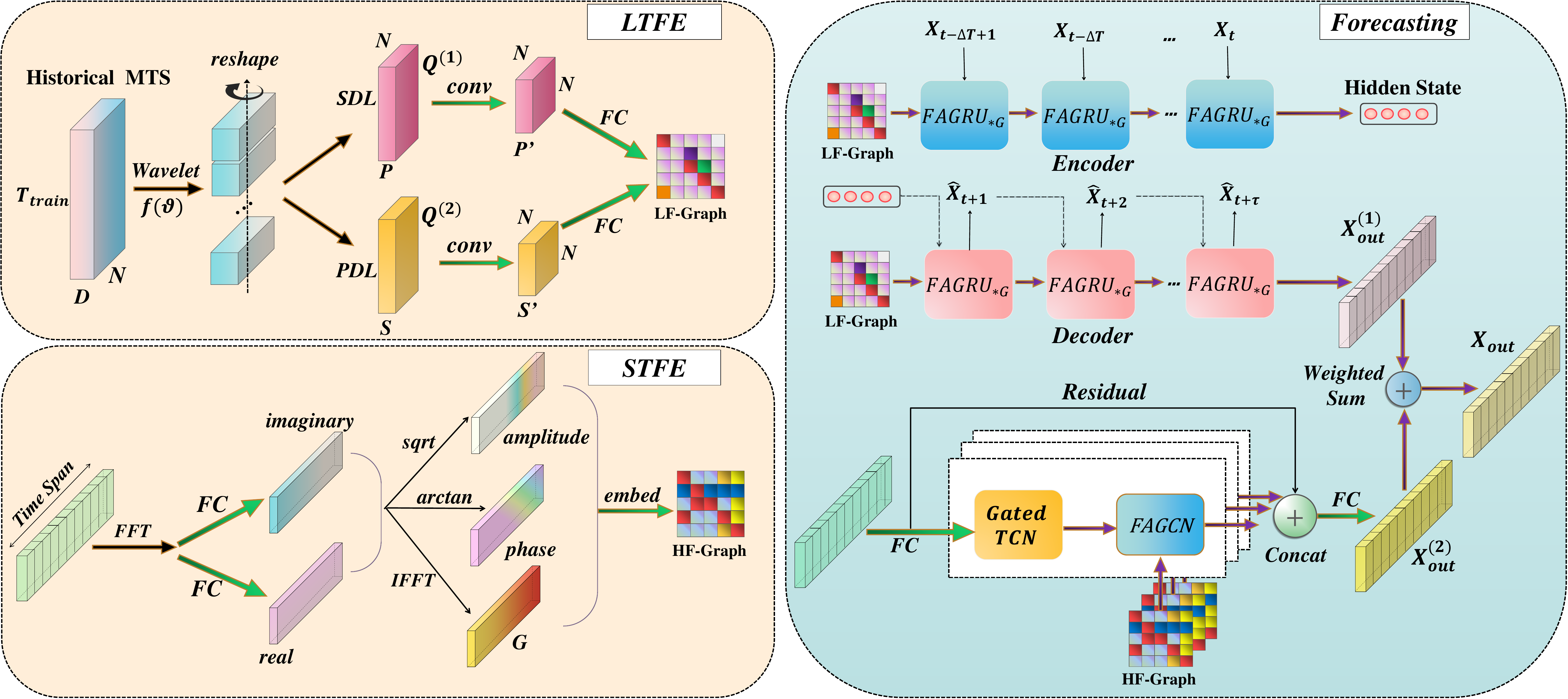}
    \caption{The framework of FCDNet. The Long-Term Frequency Extractor (LTFE) extracts low-frequency information from long historical MTS to generate the Low-Frequency Graph (LF-Graph) $A_{LF}$. For the input MTS, we apply Short-Term Frequency Extractor (STFE) from short-term input MTS to obtain High-Frequency Graph (HF-Graph) $A_{HF}$. In downstream temporal forecasting, Frequency Adaptation Graph Convolutional Gated Recurrent Unit (FAGRU) couples with $A_{LF}$ to generate the $\mathbf{X}_{out}^{(1)}$. The information of $A_{HF}$ is integrated into the Gated Temporal Convolution Network (Gated TCN) through Frequency Adaptation Graph Convolutional Networks (FAGCN) to generate $\mathbf{X}_{out}^{(2)}$. The final output is the weighted sum of $\mathbf{X}_{out}^{(1)}$ and $\mathbf{X}_{out}^{(2)}$.}
    \label{framework}
\end{figure*} 

\section{Methodology}
\subsection{Problem Formulation}
 
Let $x_t \in \mathbb{R}^{N\times D}$ denote the value of a multivariate of numbers $N$ and feature dimension $D$ at time step $t$, where $ x_t[i] \in \mathbb{R}^{D} $ denotes the $i^{th}$ variable at time step $t$. $A\in \mathbb{R}^{N\times N}$ denotes the degree of correlation between variables and $A_{i,j}\in[0,1] (i\in[1,N], j\in[1,N])$. Given the multivariate observation sequence of historical $M$ time steps $\textbf{X} = \{\textbf{x}_{t_1},\textbf{x}_{t_2},\cdots,\textbf{x}_{t_M}\}$,  our goal is to predict the future $E$-step numerical sequence $\textbf{Y} = \{\textbf{x}_{t_{M+1}},\textbf{x}_{t_{M+2}},\cdots,\textbf{x}_{t_{M+E}}\}$. Our goal is to learn a neural network $\mathcal{F}_{A}(\theta)$ for accurate forecasting.

The overall forecasting function can be written as:
\begin{align}
&[\mathbf{X}_{t_{1}:t_{M}},A]\xrightarrow{\mathcal{F}_A(\theta)}\mathbf{Y}_{t_{M+1}:t_{M+E}}.
\end{align}
The overall framework of FCDNet is present in Figure~\ref{framework}.

\subsection{Frequency-Guided Dependency Modeling}

\subsubsection{Long-Term Frequency Extractor (LTFE)}
Due to the strong volatility in MTS, it is difficult to capture stable and genuine correlations adaptively from historical MTS without prior domain knowledge. However, the low-frequency part of the time-series can reflect the general characteristics of variables and are relative invariants implied in the complex dynamics. We try to capture such low-frequency information from long-term historical time-series for modeling static and stable inter-series structures. Besides, long-term historical MTS are beneficial for resisting noise, which facilitates obtaining more robust and accurate dependencies. To this end, FCDNet adopts the training set of MTS data as long-term historical MTS. Moreover, changes in the values of different variables at cross-time can better reflect the relationships between variables. In the transportation domain, for example, the numerical changes of the sensors over time offer insights into how traffic dynamics propagate along with the network. Therefore, we first do the difference operation on the training MTS in order to reveal more moderate correlations:
\begin{equation}
\begin{aligned}
& \mathcal{D}iff(\mathbf{X}_{1},\cdots,\mathbf{X}_{T})=\{\mathbf{0}, \mathbf{X}_{2}-\mathbf{X}_{1},\cdots,\mathbf{X}_{T}-\mathbf{X}_{T-1}\}\\
&  
\triangleq \{\mathbf{\hat{X}}_{1},\mathbf{\hat{X}}_{2}, ...,\mathbf{\hat{X}}_{T}\}=\mathbf{\hat{X} }\in \mathbb{R}^{T_{train}\times N\times D}.
\end{aligned}
\end{equation}

Then, in light of the periodicity of the MTS, we set a hyper-parameter period $ P $ to segment $\mathbf{\hat{X}}$ into $ S = \lfloor T_{\text{train}}/P \rfloor$ segments, each containing time-series $\mathbf{\hat{X}}_i \in \mathbb{R}^{P \times N \times D}, i = 1,2,...,S. $ From a long-term perspective, the patterns in different time intervals can be captured in parallel by splitting the multivariate time-series into segments, reducing the number of parameters required for subsequent operations. In addition, it enables the model better capture semantic information within and between periods. After obtaining the time-series segments, we concatenate these segments to obtain a four-dimensional tensor $\mathcal{O}$:
\begin{equation}
\begin{aligned}
\mathcal{O}=[\mathbf{\hat{X}_1}\|\mathbf{\hat{X}_2}\|...\|\mathbf{\hat{X}_S}]\in \mathbb{R}^{S\times P \times N \times D}.
\end{aligned}
\end{equation}

Moreover, the real-world MTS has a low signal-noise ratio. Ignoring these noises in joint graph learning and forecasting will fail to learn the genuine dependency graph and lead to over-fitting in the forecasting. By performing the difference and segmentation operations mentioned above, noise points within a period $P$ can be more easily identified. To this end, we apply Daubchies wavelet decomposition~\cite{Wavelets} along the temporal dimension of $\mathcal{O}$, which can extract multilevel time-frequency features by decomposing the temporal dimension of $\mathcal{O}$ as low and high-frequency sub-series level by level~\cite{mallat}. Assuming $L-1$ is the number of decomposition levels, $Dec$ as the wavelet decomposition, $Rec$ as the wavelet reconstruct, and $\textbf{Db}$ as the Daubchies wavelet, we can get the coefficients of each level:
\begin{equation}
\begin{aligned}
\bm{coeffs}[i]=Dec(\mathcal{O}[s,:,n,d],&wavelet=\textbf{Db},level=i), \\
i\in [1,L],s \in[1,S],&n \in [1,N],d\in [1,D].   
\end{aligned}
\end{equation}
By controlling the coefficients, we filter out the high-frequency part of time-series and save the low-frequency part:
\begin{equation}
\begin{aligned}
\bm{C}[i]= {\bm{\lambda}}_{i} \odot \bm{coeffs}[i], i\in [1,L]. 
\end{aligned}
\label{Wavelet}
\end{equation}
Here, $\odot$ denotes an element-wise multiplication operator and $\bm{\lambda}_{i},i\in [1,L]$ are the control factors. High-frequency components can be effectively filtered by taking a small value in $\bm{\lambda}_{i}$. Final, the reorganized low-frequency information of original tensor $\mathcal{O}$ can be reconstructed from $\bm{C}[i]$:
\begin{equation}
\begin{aligned}
\mathcal{Z}[s,:,n,d,i]=Rec(\bm{C}[i],wave&let=\textbf{Db},level=i), \\ \mathcal{Z} \in \mathbb{R}^{S\times P \times N \times D \times L}.
\end{aligned}
\end{equation}
We combine the frequency dimension and feature dimension to obtain a four-dimensional tensor $\hat{\mathcal{Z}} \in \mathbb{R}^{S\times P \times N \times D \cdot L}$.

To further emphasize the uniqueness of different time segments and the particular dynamic characteristics of variables, we apply the ST-Norm across the dimension $S$ and $P$ to obtain two four-dimensional tensors:
\begin{equation}
\begin{aligned}
&\mathcal{\hat{Z}}^{(1)}=\mathbf{STN}(\hat{\mathcal{Z}}) \in \mathbb{R}^{S\times P \times N \times D \cdot L},\\
&\mathcal{\hat{Z}}^{(2)}=\mathbf{STN}(\mathcal{\hat{Z}}^{T}) \in \mathbb{R}^{P\times S \times N \times D \cdot L}.
\end{aligned}
\end{equation}

For the final parameterization of the low-frequency structure $A_{LF}$, we use a feature extractor to yield a feature representation for $A_{LF}$. We opt for a concise architecture to implement this. First of all, to make the dimensions match the subsequent 1D convolutional operations and perceive historical MTS from two different temporal perspectives, we reshape $\mathcal{\hat{Z}}^{(1)}$ into $\mathcal{Q}^{(1)}\in \mathbb{R}^{N \times S \cdot D \cdot L \times P}$ and $\mathcal{\hat{Z}}^{(2)}$ into $\mathcal{Q}^{(2)}\in \mathbb{R}^{N \times P \cdot D \cdot L \times S}$.  Then, we use 1D convolution $Conv$ along the transformed dimension and three fully connected layers to transform the three-dimensional tensor $\mathcal{Q}$ to obtain two graphs:
\begin{equation}
\begin{aligned}
A^{(i)} = \chi (FC_{1}^{(i)}(\delta (FC_{2}^{(i)}(\delta (FC_{3}^{(i)}(\delta(Conv^{(i)}(\mathcal{Q}^{(i)}))))),
\end{aligned}
\end{equation}
\begin{equation}
\begin{aligned}
A^{(i)} \in \mathbb{R}^{N\times N}
,(i=1,2)
\end{aligned}
\end{equation}
where $\delta$ represents the ReLU activation function, and $\chi$ represents the smooth sparse unit~\cite{my_work} to normalize the value of the graph matrix to be ${A}^{(i)}_{ij} \in [0,1]$, $FC_{j}^{(i)}, (j=1,2,3,i=1,2)$ are six fully connected layers. ${A}^{(1)}$ and ${A}^{(2)}$ harvest dependency information from different time perspectives. The two graphs are fused in the form of a weighted sum to obtain the low-frequency graph $A_{LF}$, where $\beta$ is a learnable parameter:
\begin{equation}
\begin{aligned}
A_{LF} = \beta A^{(1)} + (1-\beta) A^{(2)}.
\end{aligned}
\label{beta}
\end{equation}

\subsubsection{Short-Term Frequency Extractor (STFE)} 
LTFE aims to extract low-frequency information implied in long-term historical time-series for constructing a long-term stable graph structure. In addition to the genuine long-term stable correlations, the correlations between variables show different dynamics at different short-time spans. The high-changing correlations are essential for uncovering the evolution of dynamic systems and are non-neglectable in dependency modeling. For example, in the transportation domain, in addition to the inherent road network topology, different sensors will have different correlations in different periods. As a complement to the Long-Term Frequency Extractor (LTFE), we employ the Short-Term Frequency Extractor (STFE) to generate a representation of the high-frequency graph denoted as $A_{HF}$ based on the dynamic input MTS. For each input $\mathbf{X}_{in} \in \mathbb{R}^{B\times T_{in} \times N \times D}$, where $B$ denotes the batch size and $T_{in}$ denotes the length of the input sequence, we first reshape $\mathbf{X}_{in}$ into $\hat{\mathbf{X}}_{in} \in \mathbb{R}^{N \times T_{in} \times B \cdot D}$ for subsequent transform operations. Flattening the samples in both the batch and feature dimensions during the Fast Fourier Transform (FFT) is essential due to the consistent graph structure utilized by STFE when processing a batch. Maintaining this uniform graph structure throughout the batch calculation is necessary. Otherwise, generating a distinct graph structure for each sample in the batch 
will significantly escalate computational demands during subsequent calculations. Therefore, by flattening the samples in the batch and feature dimensions during FFT, STFE optimizes computational efficiency and ensures a streamlined and effective data analysis. Specifically, STFE first utilizes Fast Fourier Transform (FFT) to project $\hat{\mathbf{X}}_{in}$ into the frequency domain:
\begin{equation}
\begin{aligned}
\hat{\mathbf{X}}_{e}=FFT(\mathbf{X}_{in})\in \mathbb{R}^{N \times T_{in} \times B \cdot D}.
\end{aligned}
\end{equation}
FFT can favor STFE in learning the representations of the input time-series on the trigonometric basis in the frequency domain, which captures the repeated patterns in the periodic data or the auto-correlation features among different timestamps. Specifically, the output of FFT has real part ($\hat{\mathbf{X}}_{e}^{r}$) and imaginary part ($\hat{\mathbf{X}}_{e}^{i}$), which are processed by the linear operators with different parameters in parallel. The operations can be formulated as:
\begin{equation}
V^{*}_{u}= (\hat{\mathbf{X}}_{e}^{*})(W_{u})\in \mathbb{R}^{N \times T_{in} \times F},*\in \{r,i\}
\end{equation}
where $W_{u} \in \mathbb{R}^{B\cdot D \times F}$ is a learnable weight matrix for dimension reduction and capturing short-term frequency information across the batch and feature dimensions, $F$ is a hypermeter denotes the transformed dimension. It is well known that complex numbers can be uniquely represented by their amplitude and phase. To better encode the high-frequency graph $A_{HF}$, we represent the amplitude $\widetilde{\bf{A}}_m$ and the phase $\widetilde{\bf{S}}$ as:
\begin{equation}
\begin{aligned}
& \widetilde{\bf{A}}_m = \sqrt{(V_{u}^{r})^{2}+(V_{u}^{i})^{2}} \in \mathbb{R}^{N\times T_{in} \times F}, 
\\
& \widetilde{\bf{S}} = arctan(\frac{V_{u}^{r}}{V_{u}^{i}}) \in [-\frac{\pi}{2},\frac{\pi}{2}]^{N\times T_{in} \times F},
\end{aligned}
\end{equation}
where $arctan(\cdot)$ is an element-wise inverse tangent function. Later, we apply IFFT to get the output:
\begin{equation}
\begin{aligned}
\mathbf{G}=IFFT(V^{r}_{u}+iV^{i}_{u})\in \mathbb{R}^{T_{in} \times N \times F}.
\end{aligned}
\end{equation}
The final $A_{HF}$ is obtained by applying fully connected layers to fuse the embedding $\mathbf{G}$, $\widetilde{\bf{A}}_m$, and $\widetilde{\bf{S}}$:
\begin{equation}
\begin{aligned}
A_{HF}=\chi(W_{T}((\mathbf{G}W_{G}+\widetilde{\bf{A}}_m W_{m} + \widetilde{\bf{S}} W_{S})))\in \mathbb{R}^{N \times N},
\end{aligned}
\label{valueF}
\end{equation}
where $W_{G},W_{m},W_{S} \in \mathbb{R}^{F\times N}$, $W_{T} \in \mathbb{R}^{T_{in}}$ are learnable weight matrices for dimension transformation and feature extraction.
\subsection{Downstream Temporal Forecasting}
In addition to the representation learning of graph structures, designing an appropriate method to integrate dependency information into downstream temporal forecasting also plays a critical role in decipting dynamic systems. Specifically, we couple different downstream forecasting units for extracted $A_{LF}$ and $A_{HF}$ structures according to the characteristics of low- and high-frequency information.
\subsubsection{Frequency Adaptation Graph Convolutional Gated Recurrent Unit}
Traditional graph neural networks such as GCN~\cite{GCN} and APPNP~\cite{APPNP} are low-pass filters, which mainly retain the commonality of variables and ignore delicate learning of differences. In this regard, in our work, we adopt the graph filters of FAGCN~\cite{FAGCN} to adaptively aggregate low-frequency and high-frequency graph signals along the process of time-series message passing. FAGCN designs a low-pass filter $\mathcal{F}_{L}$ and a high-pass filter $\mathcal{F}_{H}$ to decompose the low-frequency and high-frequency parts from the graph adjacency matrix:
\begin{equation}
\begin{aligned}
&\mathcal{F}_{L} = \varepsilon I + D^{-\frac{1}{2}} A_{LF} D^{-\frac{1}{2}},
\\
&\mathcal{F}_{H} = \varepsilon I - D^{-\frac{1}{2}} A_{LF} D^{-\frac{1}{2}},
\\
&\widehat{A_{LF}} = \gamma \mathcal{F}_{L} + (1-\gamma) \mathcal{F}_{H}, 
\end{aligned}
\label{gamma}
\end{equation}
where $\gamma$ is a learnable parameter. Inspired by~\cite{DCRNN,AGCRN}, we combine FAGCN with Gated Recurrent Units (GRU)~\cite{GRU} and propose Frequency Adaptation Graph Convolutional Gated Recurrent Unit (FAGRU). Specifically, we use two-step graph convolution $\star$, which is defined as:
\begin{equation}
	W^Q_{\mathcal{G}} \star\mathbf{X}_{in}=\sum\nolimits_{k=0}^{K=2}((\widehat{A_{LF}})^{k}\mathbf{X}_{in}W^{Q}_{k}),
\end{equation}
where $W^Q_{k}, (k=0,1,2)$ are trainable parameters. 

Then, we leverage the graph convolution with Gated Recurrent Units (GRU) to model the temporal dependency:

\begin{equation}
	\begin{aligned}
		&R_t\,=\sigma(W_{\mathcal{G}}^R\star(\mathbf{X}_t||H_{t-1})+b_R),\\
		&C_t\,=tanh(W_{\mathcal{G}}^C\star(\mathbf{X}_t||R_t \odot H_{t-1})+b_C),\\
		&U_t\,=\sigma(W_{\mathcal{G}}^U\star(\mathbf{X}_t||H_{t-1})+b_U),\\
		&H_t=U_t \odot H_{t-1}+(1-U_t) \odot C_t,
	\end{aligned}
\end{equation}

where $\mathbf{X}_{t},H_t$ denote the input and output of at time $t$, $R_{t}, U_{t}$ are reset gate and update gate at time $t$ respectively responsible for catching irrelevant information to forget and the part of the past state to move forward. $||$ is concatenation along the feature dimension and $\odot$ represents the element-wise product, and $b_R$, $b_C$, $b_U$ are model parameters. The time-series output of FAGRU is denoted as $\mathbf{X}_{out}^{(1)}$:
\begin{equation}
\mathbf{X}_{out}^{(1)} = [H_{1}||H_{2}||...||H_{T_{out}}].
\end{equation}

\subsubsection{Frequency Adaptation Graph WaveNet}
Inspired by~\cite{GWN}, to capture the volatility patterns and underlying dependency implicit in short time spans, we augment fine-grained convolutions with $A_{HF}$ for joint modeling. Specifically, unlike standard 1D convolutions, the dilated causal convolution skips uniform intervals when sliding over an input sequence. Temporal Convolution Network (TCN) flexibly utilizes this property and stacks convolution layers with different dilation rates. Therefore, the receptive fields of TCN can increase exponentially with linear growth in parameters. Gated TCN consists of two TCNs and applies gating mechanisms to control information flow. Specifically, FAGWN receives the linear transform $FC_{in}$ of input $\mathbf{X}_{in}$ and forms an information bottleneck as:
\begin{equation}
\begin{aligned}
\hat{\mathbf{X}}_{in}=tanh(\mathbf{\Theta}_{1}\bigstar FC_{in}(\mathbf{X}_{in}))\odot \sigma(\mathbf{\Theta}_{2}\bigstar FC_{in}(\mathbf{X}_{in})),
\end{aligned}
\end{equation}
where $\sigma$ denotes sigmoid function, $\bigstar$ denotes the dilated convolution operation and $\mathbf{\Theta}_{1}, \mathbf{\Theta}_{2}$ denote the learnable parameters of convolution filters. Later, we employ the graph filters from FAGCN to encode different graph signals from high-frequency graph structure $A_{HF}$ and integrate it into the time-series representation $\mathbf{Z}$:
\begin{equation}
\begin{aligned}
&\mathcal{A}_{L} = \varepsilon I + D^{-\frac{1}{2}} A_{HF} D^{-\frac{1}{2}},
\\
&\mathcal{A}_{H} = \varepsilon I - D^{-\frac{1}{2}} A_{HF} D^{-\frac{1}{2}},
\\
\mathbf{Z} = \sum_{k=0}^{K_{p}} (\mathcal{A}_{L}^{k}&\hat{\mathbf{X}}_{in}W_{k1} + \mathcal{A}_{H}^{k}\hat{\mathbf{X}}_{in}W_{k2})+FC_{in}(\mathbf{X}_{in}),
\end{aligned}
\end{equation}
where $A_{\nu}^{k},\nu\in \{L,H\}$ represent different-level power series of the graph matrices, $W_{k1},W_{k2},(k=0,...,K_p)$ are learnable weight matrices. We stack multiple layers of Gated TCN and FAGCN and perform computation iteratively, where the input of the next layer is the output of the current layer. The compact representation $\mathbf{X}_{out}^{(2)}$ is obtained by the linear transform of multiple outputs:
\begin{equation}
\mathbf{X}_{out}^{(2)} = \delta(FC_{1}^{(3)}(\delta(FC_{2}^{(3)}[\mathbf{Z}_{1}||\mathbf{Z}_{2}||...||\mathbf{Z}_{K}]))),
\end{equation}
where $FC_{1}^{(3)},FC_{2}^{(3)}$ are two fully connected layers, $K$ is the number of stacked layers. The final output of FCDNet is:
\begin{equation}
\begin{aligned}
\mathbf{X}_{out} = (\mathbf{X}_{out}^{(1)}+\eta \mathbf{X}_{out}^{(2)})/(1+\eta) \in \mathbb{R}^{B\times T_{out} \times N \times D_{out}},
\end{aligned}
\label{outputs}
\end{equation}
where $\eta\ge 0$ is a learnable parameter, $T_{out}$ denotes the output length, $D_{out}$ denotes the output feature dimension.
\begin{table}
\centering
\resizebox{\linewidth}{!}{
\begin{tabular}{l|ccccc}  
				\toprule         
				Datasets&\#Samples&\#Nodes&Sample Rate& \ Input length &\ Output length \cr
				\midrule 
				PEMS03&26,208&358&5 minutes&12&12 \cr
    			PEMS04&16,992&307&5 minutes&12&12 \cr
    			PEMS07&28,224&883&5 minutes&12&12 \cr
    			PEMS08&17,856&170&5 minutes&12&12 \cr
                Solar-Energy&52,560&137&10 minutes&12&12 \cr
                Ashare&937&3186&1 day&12&3\cr
				\bottomrule 
			\end{tabular}
}

\caption{Dataset statistics.}
\label{tab:dataset}
\end{table}
\section{Experiments}

We verify FCDNet on six MTS datasets, including PEMS03, PEMS04, PEMS07, and PEMS08, which were collected by Caltrans Performance Measurement System (PEMS) ~\cite{dataset}, Solar-Energy contains the solar power output from 137 PV plants in Alabama State in 2007, and Ashare contains the price indicators of stocks in Shanghai and Shenzhen exchange markets from January 2019 to November 2022 with forecasting target as each stock's closing price. 
Detailed data statistics are provided in Table~\ref{tab:dataset}. Z-score normalization is applied to the inputs. To be consistent with most modern methods, we split the PEMS datasets into training, validation, and test sets in a 6:2:2 ratio. For Solar-Energy and Ashare datasets, the split ratio is 7:1:2. All of the methods are evaluated with three metrics: mean absolute error (MAE), root mean square error (RMSE), and mean absolute percentage error (MAPE). Due to the non-uniform distribution of the solar power output of PV plants in spatial and temporal domains, there are many zeros in Solar-Energy. Hence, we only adopt MAE and RMSE for this dataset. We demonstrate the significant shortcomings of the current forecasting model in dealing with low signal-to-noise ratio MTS using the Ashare dataset.

\subsection{Baselines}
We compare FCDNet with the following models: 

(1) \textbf{VAR} Vector Auto-Regression~\cite{VAR}. 

(2)~\textbf{FC-LSTM}~\cite{FC-LSTM}, which is a recurrent neural network with fully connected LSTM hidden units; 

(3)~\textbf{DCRNN}~\cite{DCRNN}, which integrates graph convolution into a gated recurrent unit; 

(4)~\textbf{GWN}~\cite{GWN}, which integrates diffusion graph convolutions with 1D dilated convolutions;

(5)~\textbf{MTGNN}~\cite{MTGNN}, which uses external features to automatically generate graphs coupled with dilated inception layers for MTS forecasting;

(6)~\textbf{ASTGCN}~\cite{ASTGCN}, which introduces a spatial-temporal attention mechanism in the model;

(7)~\textbf{STG-NCDE}~\cite{STG-NCDE}, which combines two neural controlled differential equations for the spatial-temporal processing;

(8)~\textbf{STGODE}~\cite{STGODE}, which applies continuous graph neural network with ordinary differential equations for multivariate time-series forecasting;

(9)~\textbf{Z-GCNETs}~\cite{Z-GCNETs}, which introduces the concept of zigzag persistence into a time-aware graph convolutional network for time-series prediction;

(10)~\textbf{AGCRN}~\cite{AGCRN}, which exploits learnable embedding factorization of nodes in graph convolution;

(11)~\textbf{DSTAGNN}~\cite{DSTAGNN}, which represents dynamic spatial relevance among nodes with an improved multi-head attention mechanism and acquires the range of dynamic temporal dependency via multi-scale gated convolution;

(12)~\textbf{FEDformer}~\cite{FedFormer}, which  combines transformer with the seasonal-trend decomposition method;

(13)~\textbf{FiLM}~\cite{Film}, which applies Legendre Polynomials projections to approximate historical information and uses Fourier projection to remove noise.

(14)~\textbf{ST-Aware}~\cite{Aware}, which aims at turning spatial-temporal agnostic models into spatial-temporal aware models.

\begin{table*}
\centering
\resizebox{\textwidth}{!}{
\begin{tabular}{clccc|ccc|ccc|ccc|cc|ccc}  
				\toprule

				\multirow{2}{*}{}&\multirow{2}{*}{ Models}&
				\multicolumn{3}{c}{PEMS03}&\multicolumn{3}{c}{PEMS04}&\multicolumn{3}{c}{PEMS07}&
				\multicolumn{3}{c}{PEMS08}&\multicolumn{2}{c}{Solar-Energy}&\multicolumn{3}{c}{Ashare}\cr 
				
				\cmidrule(lr){3-19}   	
				&&MAE&RMSE&MAPE  &MAE&RMSE&MAPE&MAE&RMSE&MAPE&MAE&RMSE&MAPE&MAE&RMSE&MAE&RMSE&MAPE\cr
								\midrule
				\multirow{9}{*}{}

                &VAR&23.65&38.26&24.51$\%$&24.54&38.61&17.24$\%$&50.22&75.63&32.22$\%$&19.19&29.81&13.10$\%$&2.71&4.10&37.06&517.49&28.22$\%$\cr

				&FC-LSTM&21.33&35.11&22.33$\%$&26.24&40.49&19.30$\%$&29.96&43.94&14.34$\%$&22.20&33.06&15.02$\%$&1.32&3.38&5.94&146.86&4.21$\%$\cr
				&DCRNN&18.18&30.31&18.91$\%$&24.70&38.12&17.12$\%$&25.30&38.58&11.66$\%$&17.86&27.83&11.45$\%$&0.94&2.51&7.79&164.13&6.88$\%$\cr
				
				&GWN&19.12&32.77&18.89$\%$&24.89&39.66&17.29$\%$&26.39&41.50&11.97$\%$&18.28&30.05&12.15$\%$&1.94&3.25&5.56&147.52&3.92$\%$\cr
				
				&MTGNN&15.30&26.17&14.93$\%$&19.50&32.00&14.04$\%$&20.94&\underline{34.03}&9.10$\%$&\underline{15.31}&\underline{24.42}&10.70$\%$&1.14&3.48&13.90&136.56&7.83$\%$\cr
				
				&ASTGCN&17.69&29.66&19.40$\%$&22.93&35.22&16.56$\%$&28.05&42.57&13.92$\%$&18.61&28.16&13.08$\%$&1.01&2.74&26.24&529.50&4.55$\%$\cr
				
				&STG-NCDE&15.57&28.34&16.30$\%$&19.21&31.09&12.76$\%$&20.53&33.84&8.80$\%$&15.45&24.81&\underline{9.92$\%$}&1.24&3.00&20.65&430.74&15.60$\%$\cr
				
				&STGODE&16.50&27.84&16.69$\%$&20.84&32.82&13.77$\%$&22.59&37.54&10.14$\%$&16.81&25.97&10.62$\%$&2.37&4.04&31.13&250.60&31.80$\%$\cr
    
				&Z-GCNETs&16.64&28.15&16.39$\%$&19.50&31.61&12.78$\%$&21.77&35.17&9.15$\%$&15.76&25.11&10.01$\%$&0.90&2.54&42.88&592.48&7.22$\%$\cr
				
				&AGCRN&15.98&28.25&15.23$\%$&19.83&32.26&12.97$\%$&22.37&36.55&9.12$\%$&15.95&25.22&10.09$\%$&\underline{0.87}&\underline{2.49}&39.08&543.20&13.77$\%$\cr

                &DSTAGNN&15.57&27.21&\underline{14.68$\%$}&19.30&31.46&12.70$\%$&21.42&34.51&9.01$\%$&15.67&24.77&9.94$\%$&0.97&2.54&15.77&380.53&14.79$\%$\cr

                &FEDFormer&16.12&\underline{25.87}&16.95$\%$&20.92&32.99&15.07$\%$&23.72&37.37&10.53$\%$&17.45&27.20&11.40$\%$&1.26&2.96&5.70&102.53&5.55$\%$\cr

                &FiLM&21.43&34.60&22.54$\%$&27.89&43.73&19.73$\%$&31.73&49.43&14.79$\%$&22.33&35.47&14.70$\%$&1.76&3.93&\underline{4.35}&\underline{93.12}&\underline{3.18$\%$}\cr

                &ST-WA&\underline{15.17}&26.63&15.83$\%$&\underline{19.06}&\underline{31.02}&\underline{12.52$\%$}&\underline{20.74}&34.05&\underline{8.77$\%$}&15.41&24.62&9.94$\%$&2.56&4.34&21.65&458.59&12.95$\%$\cr

                &FCDNet&\bf{14.38}&\bf{25.30}&\bf{14.35}$\%$&\bf{18.28}&\bf{30.37}&\bf{12.39}$\%$&\bf{19.68}&\bf{32.63}&\bf{8.37}$\%$&\bf{13.93}&\bf{23.13}&\bf{9.17}$\%$&\bf{0.81}&\bf{2.38}&\bf{3.89}&\bf{82.18}&\bf{2.89}$\%$\cr

                \midrule
                &\bf{Improve.}&\bf{5.21}$\%$&\bf{2.20}$\%$&\bf{2.25}$\%$&\bf{4.09}$\%$&\bf{2.10}$\%$&\bf{1.04}$\%$&\bf{5.11}$\%$&\bf{4.11}$\%$&\bf{4.56}$\%$&\bf{9.01}$\%$&\bf{5.28}$\%$&\bf{7.56}$\%$&\bf{6.90}$\%$&\bf{4.41}$\%$&\bf{10.57}$\%$&\bf{11.75}$\%$&\bf{9.12}$\%$\cr

				\bottomrule 
			\end{tabular}
}
\caption{{Performance of FCDNet and baselines on PEMS03, PEMS04, PEMS07, PEMS08, Solar-Energy, and Ashare. Bold \& underlines depict the best \& second-best results. The improvement over SOTA is statistically significant (t-test $p<0.01$).}}
\label{tab:performance1}
\end{table*}

\subsection{Experimental Setups}
We implement our experiments on the platform PyTorch 1.10.1 using one NVIDIA GeForce RTX 3090 GPU and tune all hyperparameters on the validation data by grid search for FCDNet. For the proposed framework, the period $P$ is searched within $\{10, 20, 30, 40\}$ for Ashare and $\{180, 216, 252, 288, 324\}$ for other datasets. The decomposition level $L$ is searched within $\{2, 3, 4, 5\}$, the initial learning rate is searched within $\{1e^{-3}, 2e^{-3}, 3e^{-3}, 4e^{-3}, 5e^{-3}\}$, and feature size $F$ in~(\ref{valueF}) is searched within $\{5, 10, 15, 20\}$. In the final implementation, the hyperparameter period $P$ is set to 20 for Ashare and 288 for other datasets (i.e., segmentation in days). The hyperparameter $L$ is set to 5, which is equivalent to using the four levels daubchies decomposition. We follow the design in~\cite{FAGCN,GWN} and set $\varepsilon$ to 0.3, $K_p$ to 2, and $K$ to 4. The $\beta$ in~(\ref{beta}), $\eta$ in~(\ref{outputs}), and $\gamma$ in~(\ref{gamma}) are initialized as 0.9, 0.1, and 0.8, respectively. Mean absolute error (MAE) is taken as the training objective of FCDNet. The $F$ in~(\ref{valueF}) is set as 5 for Ashare and 10 for other datasets. Missing values are excluded both from training and testing. The initial learning rate is $3e^{-3}$ with a decay rate of 0.1 per 10 epochs, and the minimum learning rate is $3e^{-5}$. The batch size is set to 4 for Ashare and 64 for other datasets. We adopt the Adam optimizer, and the number of training epochs is set as 250. Note that some baselines require a predefined graph structure between forecast variables, which is not available in Solar-Energy and Ashare. In this regard, we construct a $k$NN graph ($k\in \{5, 10, 20\}$) as the predefined graph structure following previous methods~\cite{GTS,Pretraining}.We repeated all experiments five times and reported the average results. 
\begin{figure}
    \centering
    \includegraphics[width=0.4952\columnwidth]{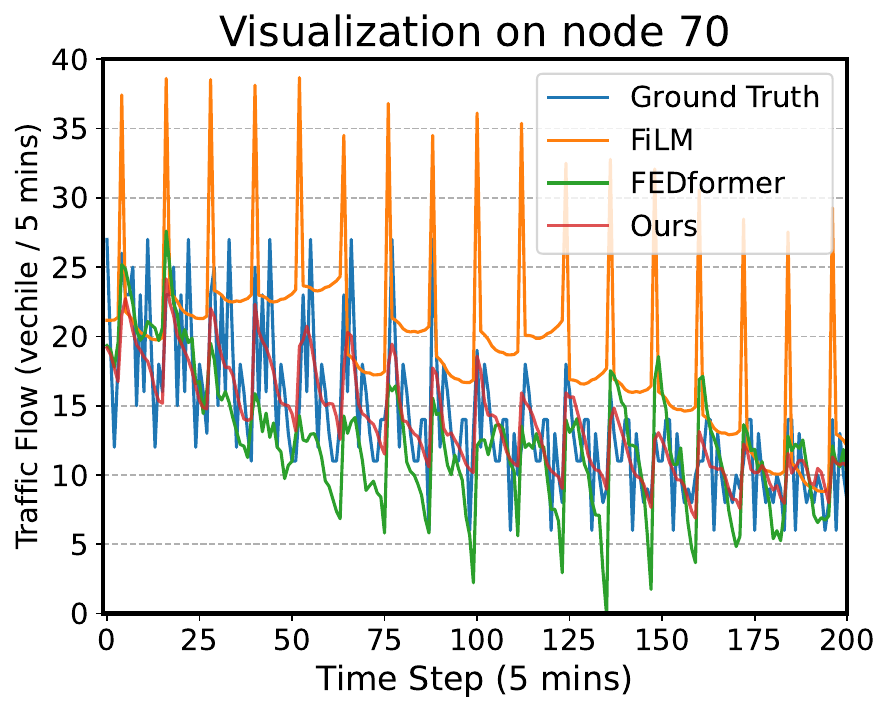}
    \includegraphics[width=0.4952\columnwidth]{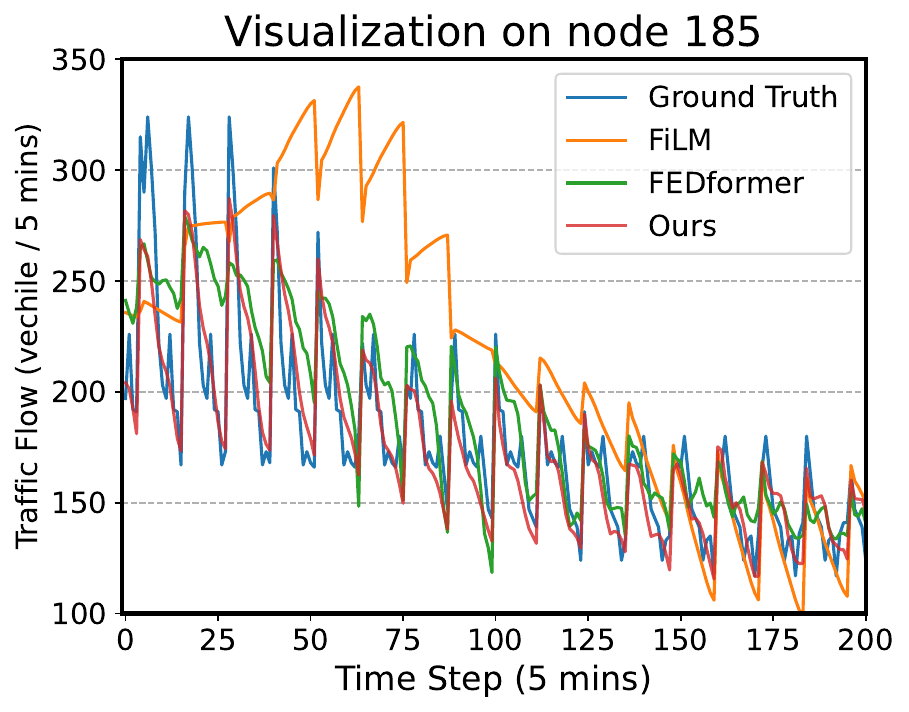}
    \caption{ Comparison of prediction curves between FEDformer, FiLM, and our FCDNet on a snapshot of the test
    data of PEMS03.}
    \label{plot}
\end{figure} 

\begin{figure}
    \centering
    \includegraphics[width=0.4952\columnwidth]{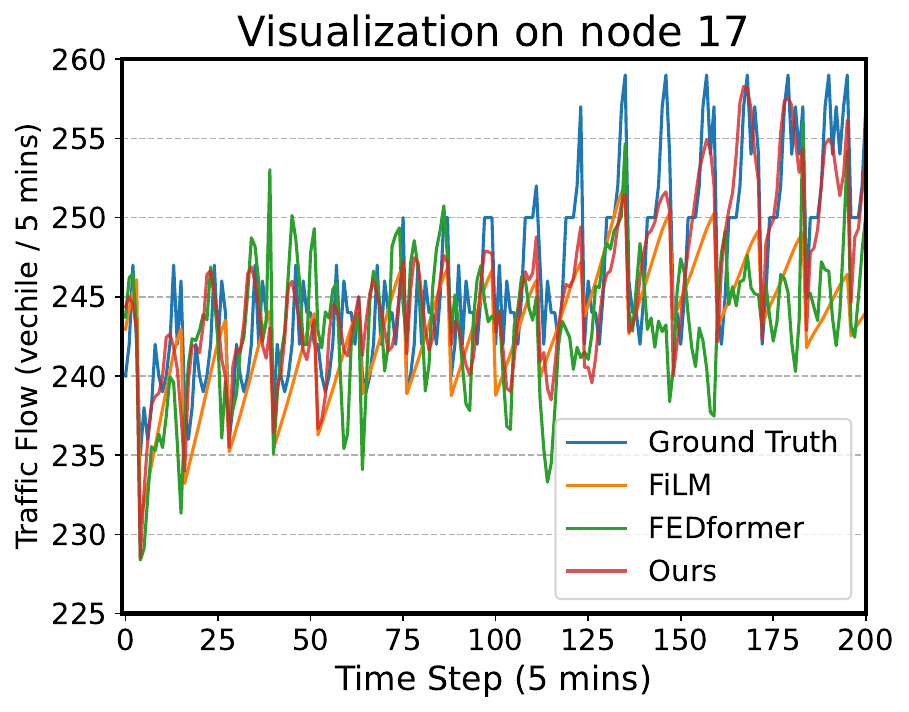}
    \includegraphics[width=0.4952\columnwidth]{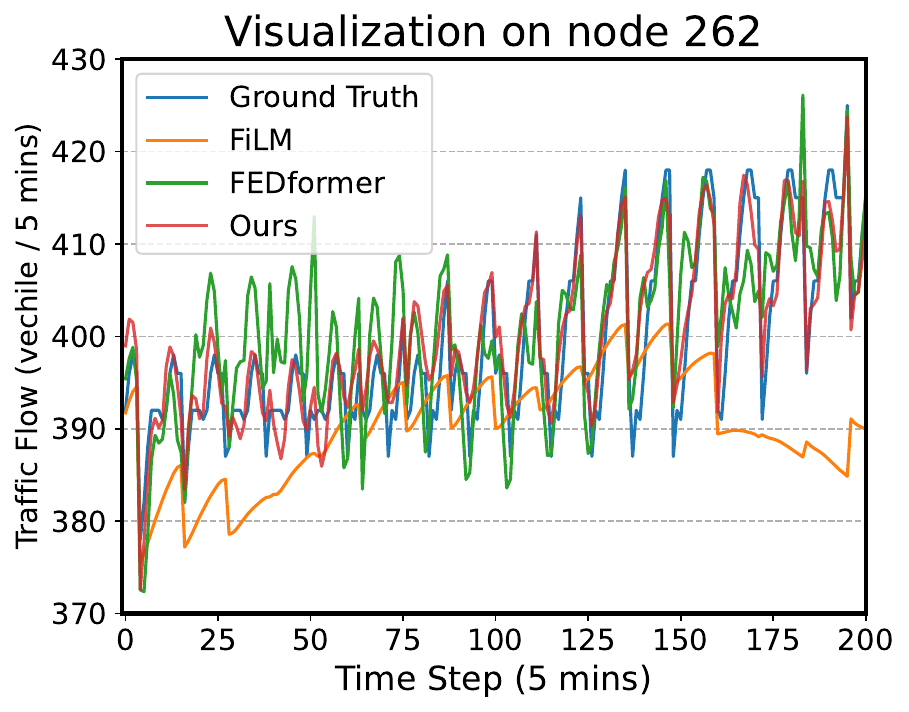}
    \caption{ Comparison of prediction curves between FEDformer, FiLM, and our FCDNet on a snapshot of the test
    data of PEMS07.}
    \label{plot}
\end{figure}

\begin{table}
\centering
\resizebox{\linewidth}{!}{
\begin{tabular}{clc}  
				\toprule
				Data&Models&Parameters \cr
				
				\midrule 
				
				\multirow{6}{*}{\shortstack{PEMS-\\04}}
				
				&MTGNN&549k\cr
				
				&AGCRN&749k\cr
    
				&STG-NCDE&2,550k\cr
    
	            &DSTAGNN&3,580k\cr

                &FEDformer&100,999k\cr

                &ST-WA&471k\cr
                
				&FCDNet&\bf{356k}\cr
				
				\bottomrule 
			\end{tabular}
\ \ \ \ \ \ \ \ \ \ \ \ \ \ 
\begin{tabular}{clc}  
				\toprule
				Data&Models&Parameters \cr
				
				\midrule 
				
				\multirow{6}{*}{\shortstack{Ashare}}
				
				&MTGNN&2,244k\cr
				
				&AGCRN&777k\cr

                &STG-NCDE&2,606k\cr
                
	        	&DSTAGNN&136,366k\cr

                &FEDformer&101,906k\cr
                
                &ST-WA&2,182k\cr
				
				&FCDNet&\bf{539k}\cr
				
				\bottomrule 
			\end{tabular}
}
\label{parameter}
\caption{Trainable parameters of different high-performing models on PEMS04 and Ashare when achieving the best results.}
\end{table}

\begin{table}
\centering
\resizebox{\linewidth}{!}{
\begin{tabular}{clcc|ccc}  
				\toprule         
			
				&\multirow{2}{*}{Models}&
				\multicolumn{2}{c}{Solar-Energy}&\multicolumn{3}{c}{Ashare}\cr 
				
				\cmidrule(lr){3-7}   	
				&&MAE&RMSE&MAE&RMSE&MAPE\cr
								\midrule
				\multirow{3}{*}{}
                
                &FCDNet&\bf{0.81}&\bf{2.38}&\bf{3.89}&\bf{82.18}&\bf{2.89}$\%$\cr

                &w/o STFE&0.85&2.47&7.29&106.52&5.32$\%$\cr

                &w/o LTFE&0.94&2.62&10.57&113.79&8.97$\%$\cr
			
				\bottomrule 
			\end{tabular}
}
\caption{{The ablation study on Solar-Energy and Ashare.}}
\label{tab:ablation}
\end{table}
 
\subsection{Experimental Results and Analysis}
Table~\ref{tab:performance1} shows the average results of FCDNet and fourteen baseline methods, from which we have several findings. \textbf{(1)} It can be seen that our model significantly outperforms other baselines across all metrics on the six datasets, illustrating the effectiveness of our model. The gained improvement of FCDNet compared with other baseline models lies in the mining and integration of long short-term time-frequency signals with complementary dependency modeling.~\textbf{(2)} On the Ashare dataset, many complex models perform poorly, but simple models such as FC-LSTM and GWN perform well. This is probably because the stock time-series shows more substantial volatility and a lower signal-to-noise ratio. Thus, the lack of distinguishing robust features from blended noise data can easily lead to serious over-fitting of the complex model. The performance of FiLM and FEDFormer on the Ashare dataset is also impressive, both focusing on extracting time series fluctuation patterns in the frequency domain.~\textbf{(3)} To further compare with two representative frequency-based models, we also conducted visualization experiments on the PEMS03 test set, as shown in Figure~\ref{plot}. FiLM performs poorly on traffic datasets, possibly due to its weak ability to capture short-term temporal information. Our model also demonstrates a stronger capability to capture multivariate temporal dynamics.~\textbf{(4)} It can be seen that some models, such as STGODE and ST-WA, are highly dependent on the prior graph and cannot perform well in the forecasting tasks without prior domain correlation knowledge. It demonstrates the importance of investigating adaptive structure modeling.~\textbf{(5)} Different from previous models, FCDNet proposes to extract correlation from multi-frequency information to discover more essential and complementary static-dynamic relationships between MTS. It can also be seen from Table 3 that the model parameters of FCDNet increase linearly with the number of forecast variables by a small factor, exhibiting better scalability and robustness.

\begin{figure}
    \centering
    \vspace{-0.3cm}
    \includegraphics[width=0.4952\columnwidth]{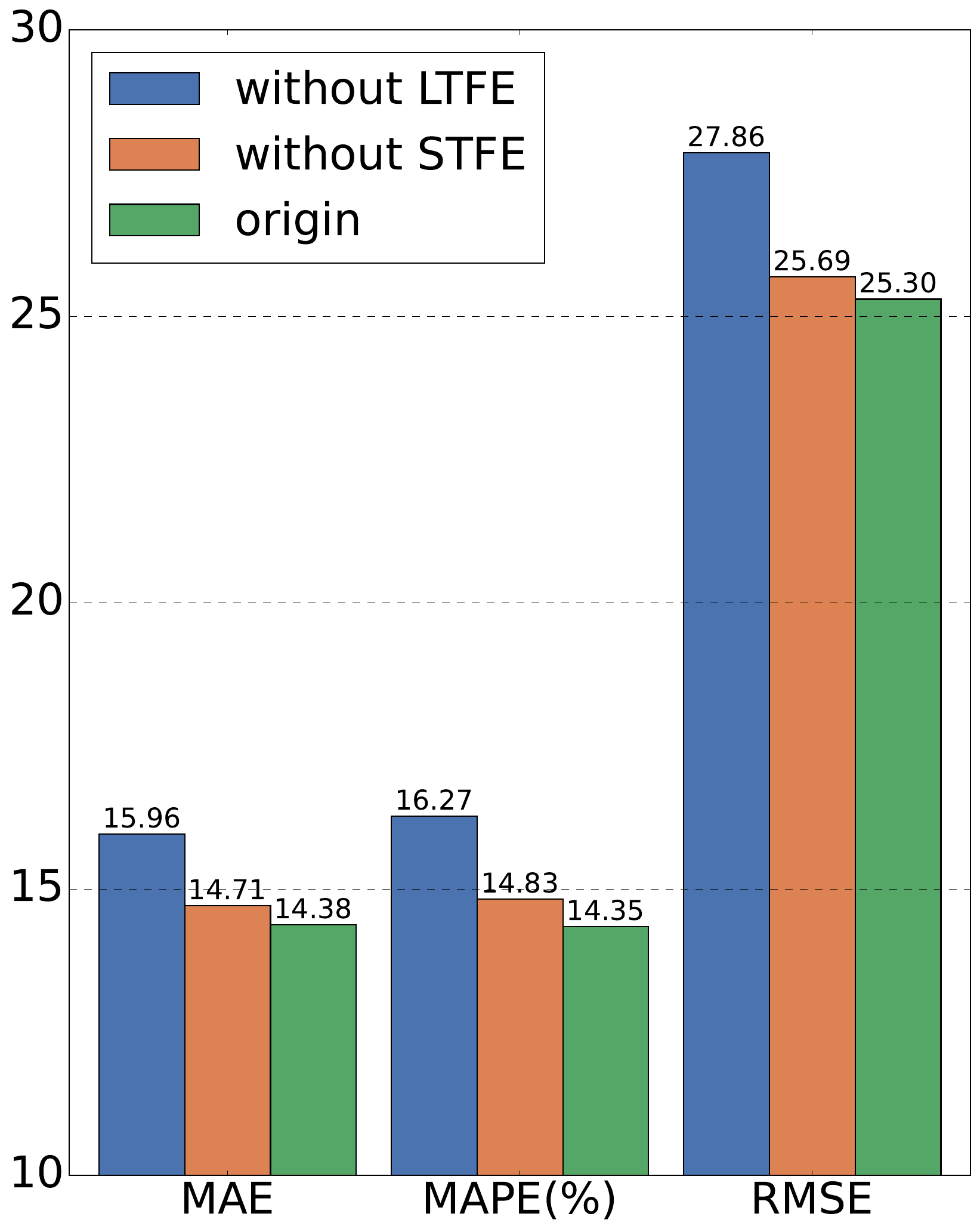}\includegraphics[width=0.4952\columnwidth]{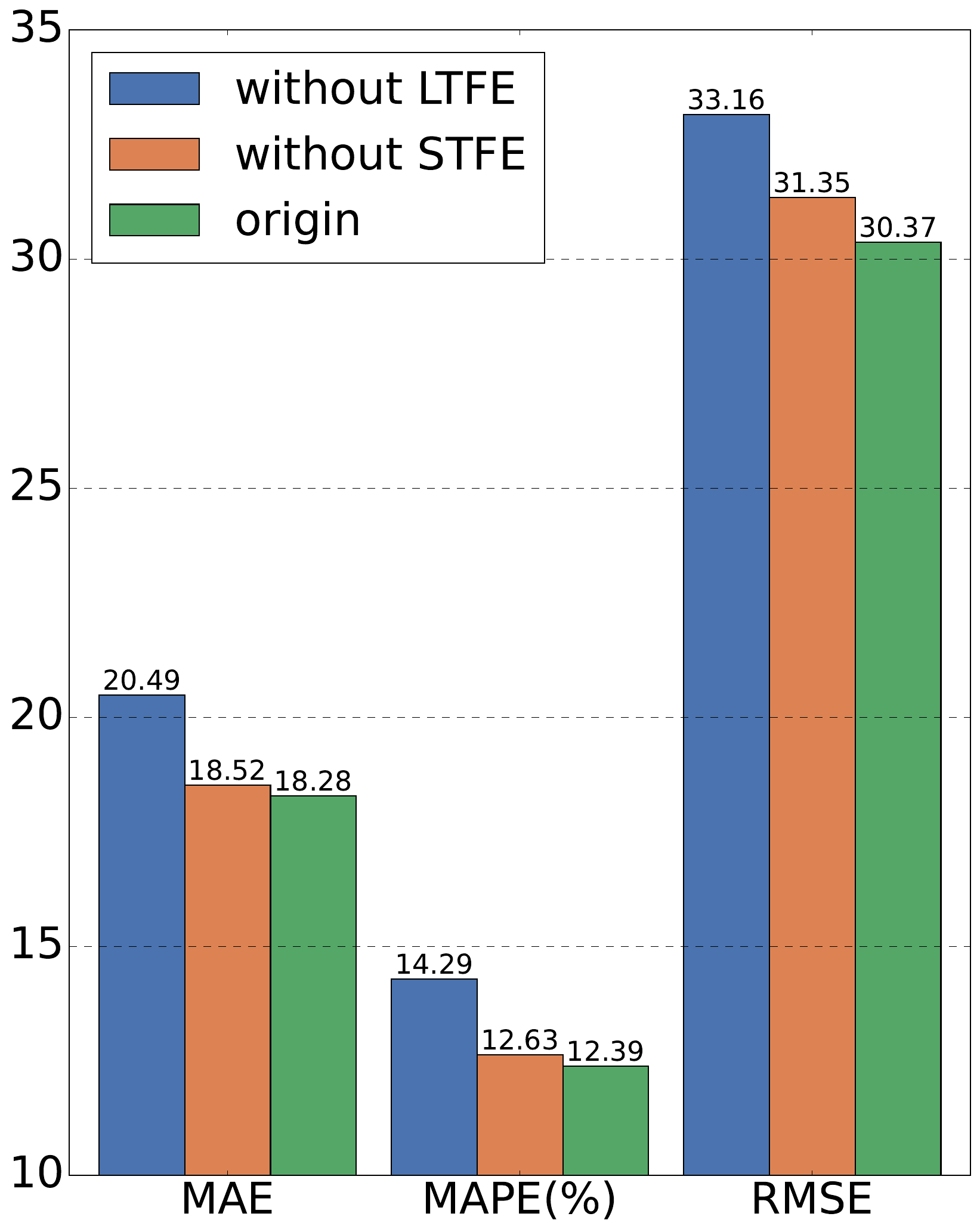}
    \includegraphics[width=0.4952\columnwidth]{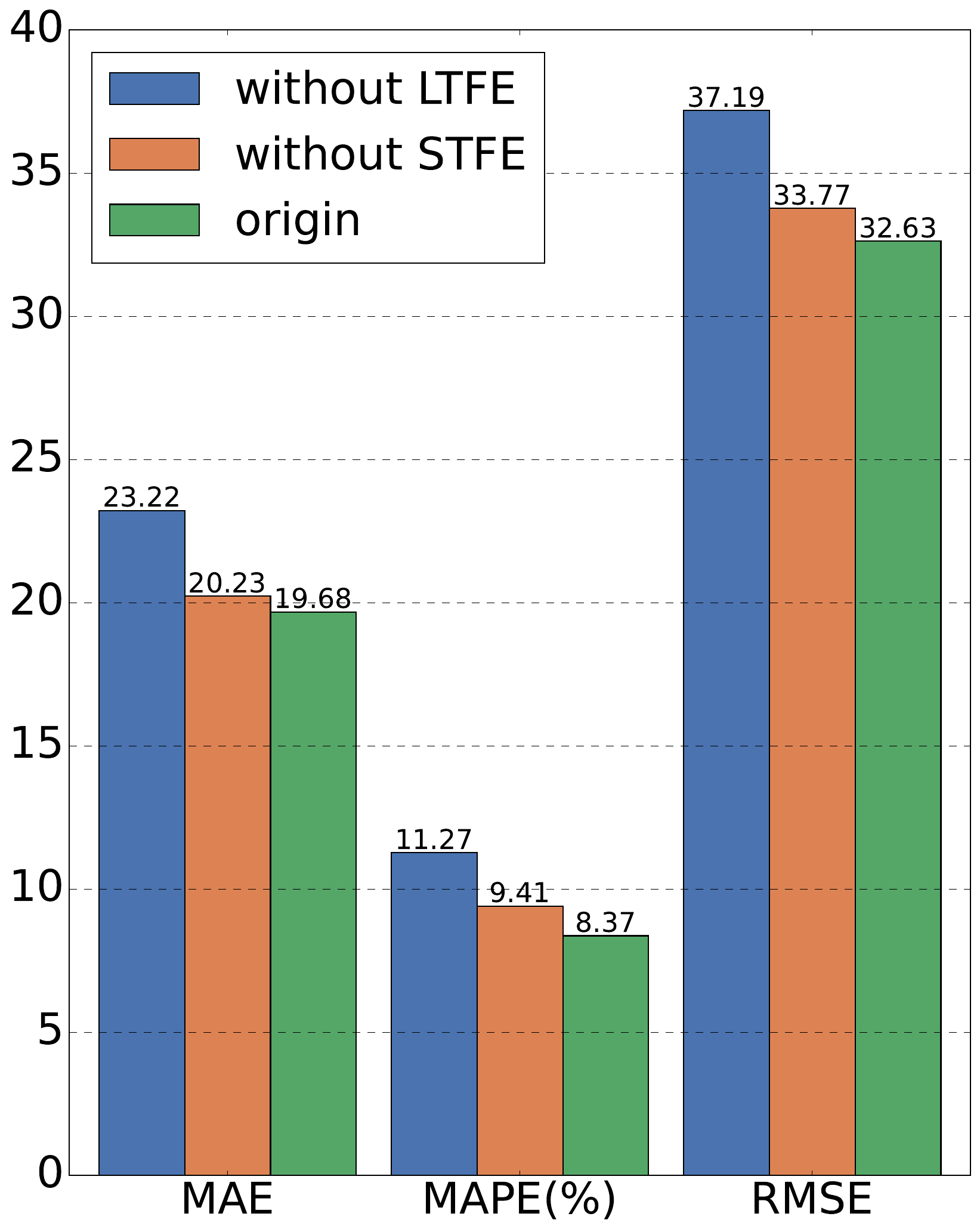}\includegraphics[width=0.4952\columnwidth]{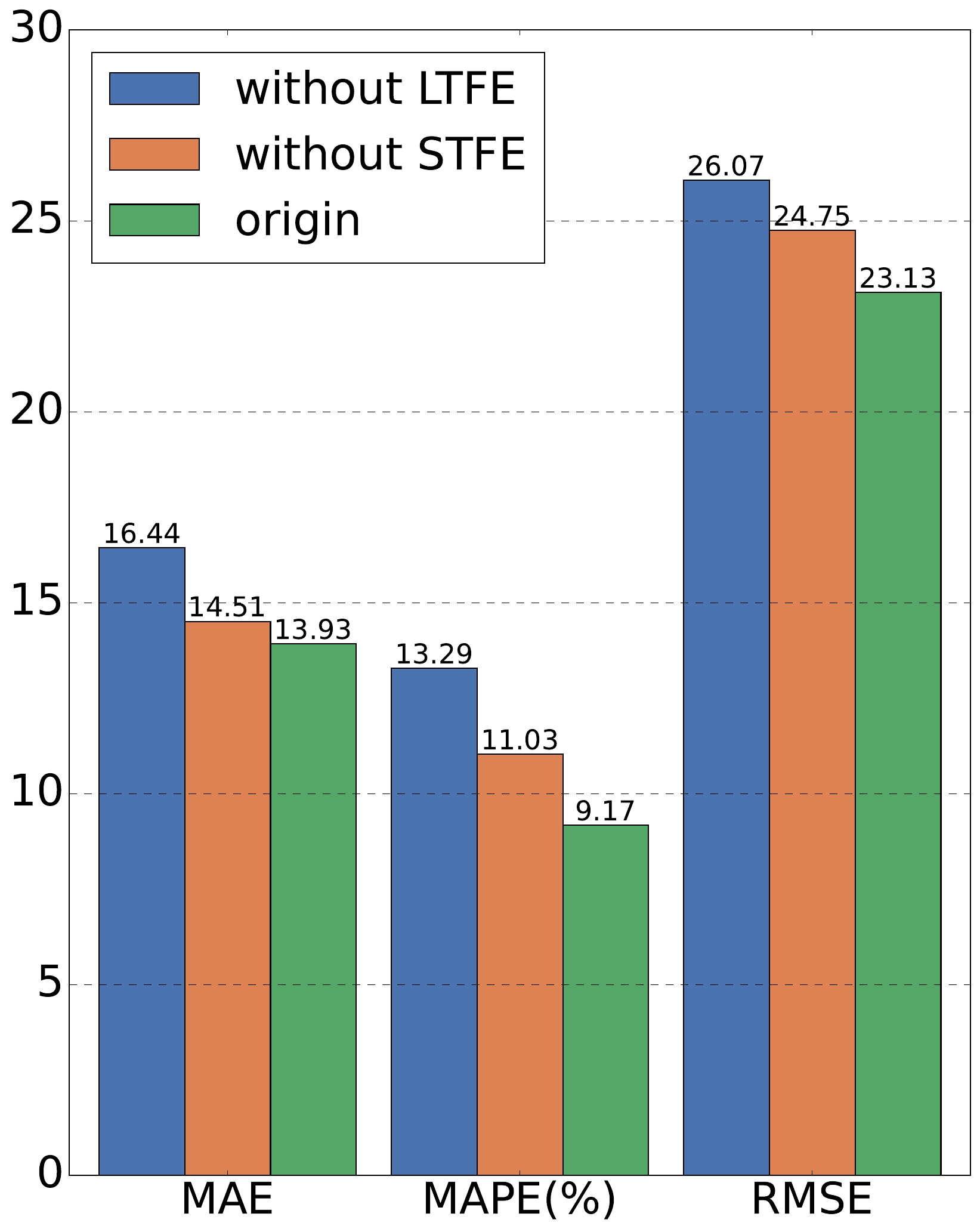}
    \caption{The ablation study on PEMS03 (upper left), PEMS04 (upper right), PEMS07 (lower left), and PEMS08 (lower right).}
    \label{ablation}
\end{figure} 

\subsection{Ablation Study}
To further verify the effectiveness of LTFE and STFE proposed by FCDNet, we replace LTFE or STFE with a randomly initialized matrix using low-rank approximation and optimize them in the training process through an end-to-end approach. It is worth noting that many models, such as MTGNN, GWN, and AGCRN, take this form in structural modeling. The ablation results are shown in Figure~\ref{ablation} and Table~\ref{tab:ablation}. LTFE assists the model in grasping the low-frequency information implicit in the long historical MTS, and STFE helps the model capture the high-changing information from the new input MTS. It can be seen that long-term stable dependencies and short-term immediate interactions are complementary, helping the model achieve the best results. From Formula~(\ref{outputs}), we have allocated more weight proportion to LTFE, which is the reason that removing the LTFE module in the ablation study has a greater impact on overall prediction than removing the STFE module. This design also contains a simple curriculum learning idea i.e., let the model capture low-frequency information first and then adapt to high-frequency information, which is likely to be subject to more volatility. In particular, we can see that LTFE can effectively mine interdependencies in the long historical MTS and boosts the forecasting performance significantly. STFE and LTFE complement each other, which can more flexibly capture temporal and structural characteristics that appear in short-term segments and further optimize the model's capability for forecasting.

\begin{figure}
    \centering
    \vspace{-0.3cm}
    \includegraphics[width=0.4952\columnwidth]{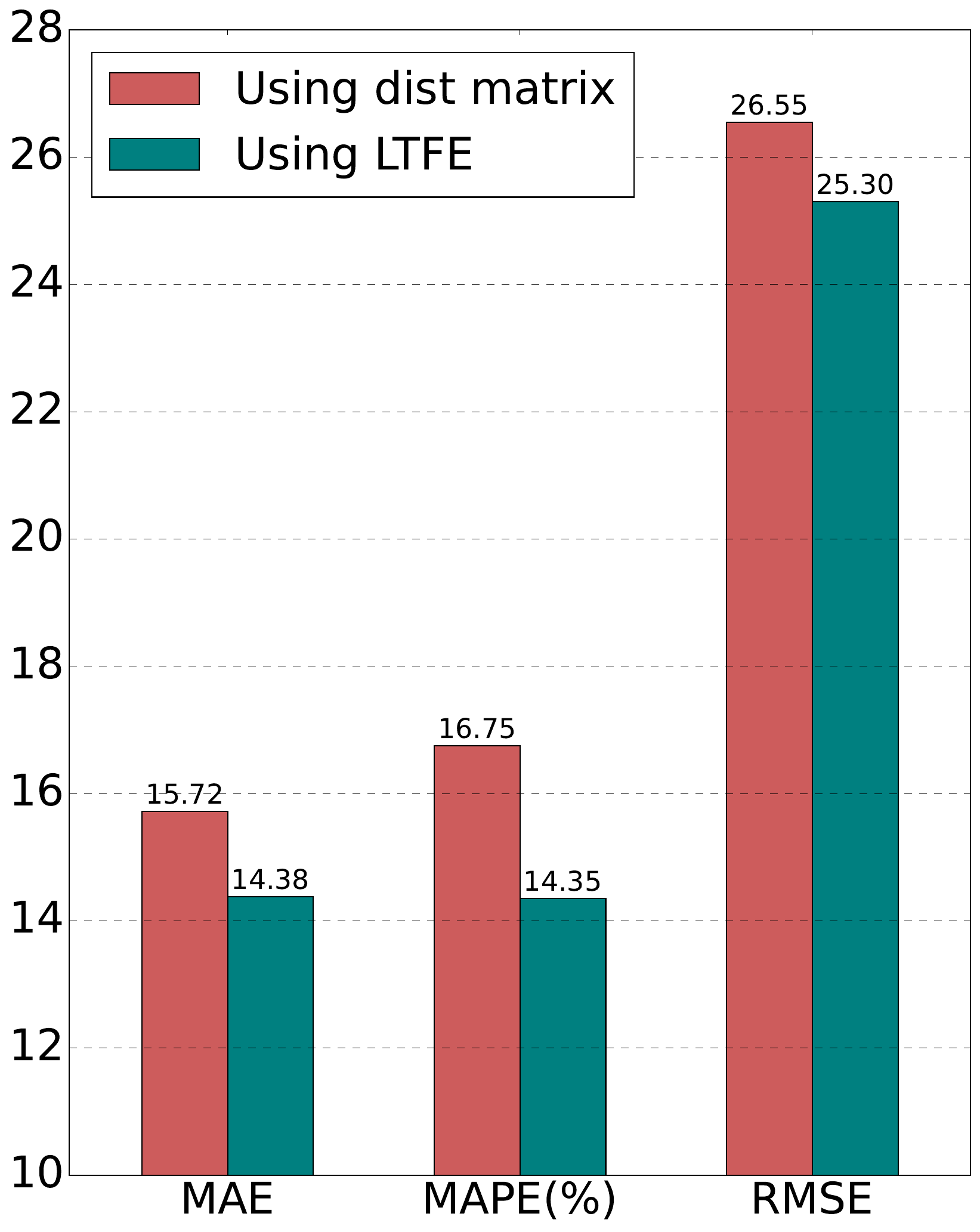}\includegraphics[width=0.4952\columnwidth]{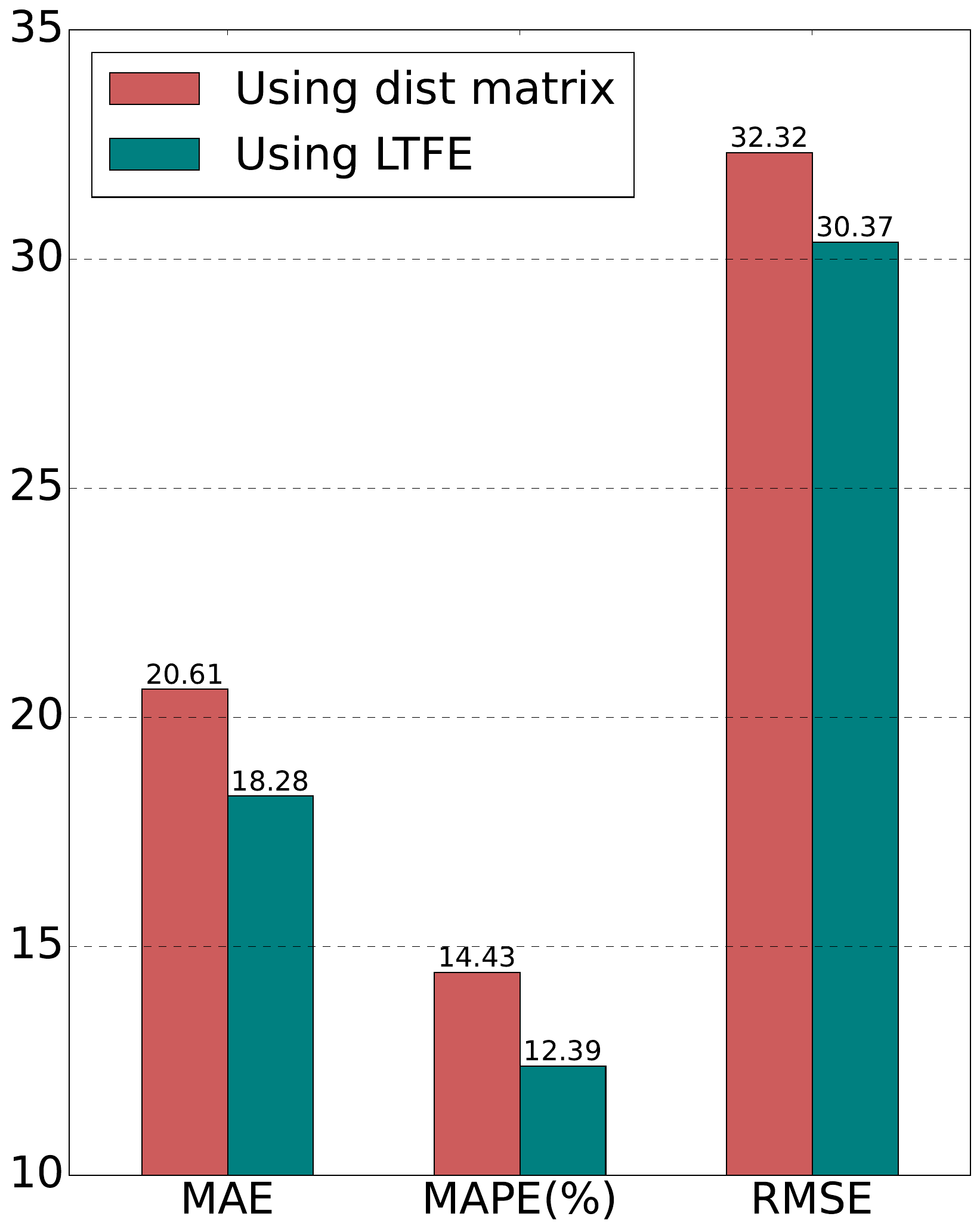}
    \includegraphics[width=0.4952\columnwidth]{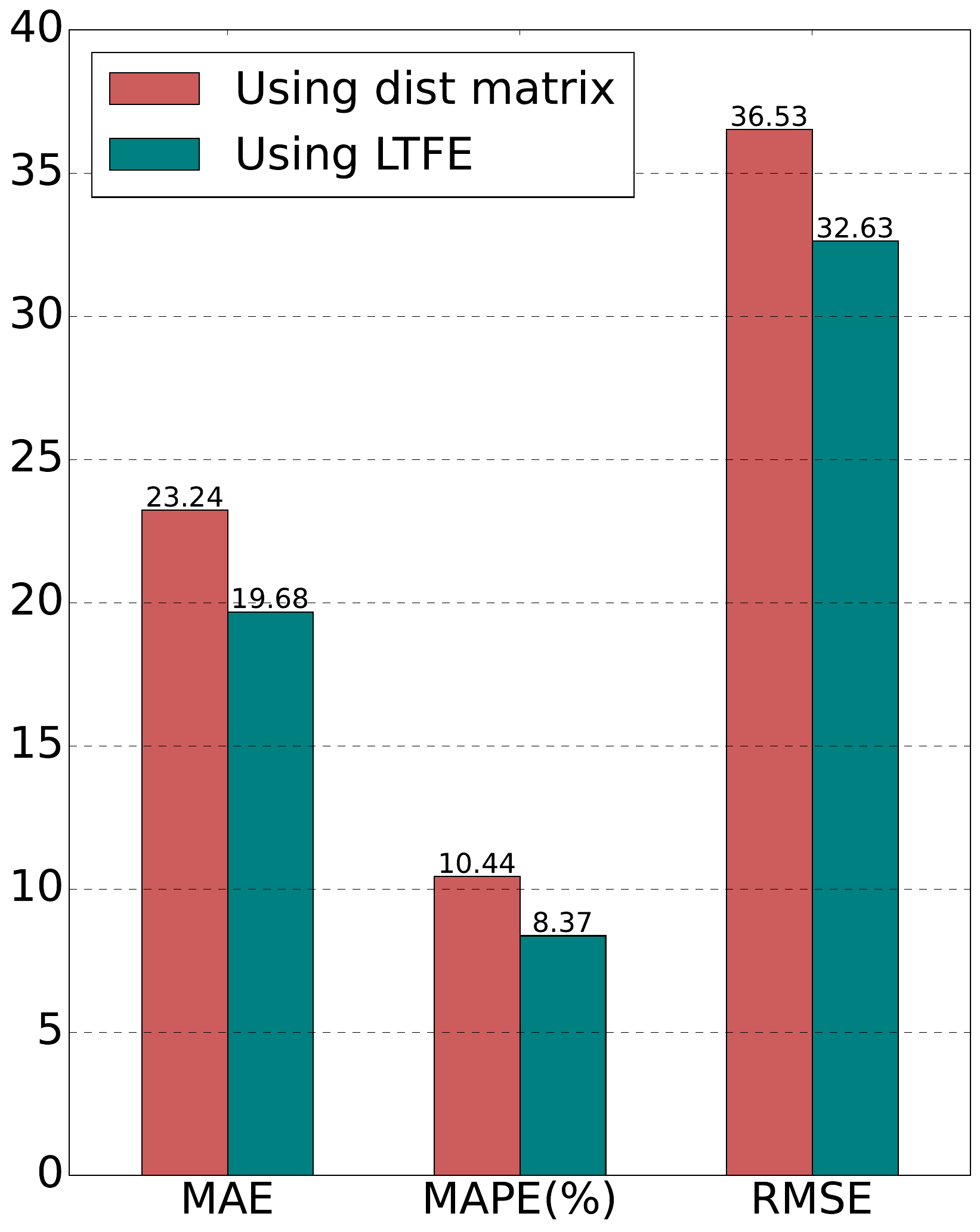}\includegraphics[width=0.4952\columnwidth]{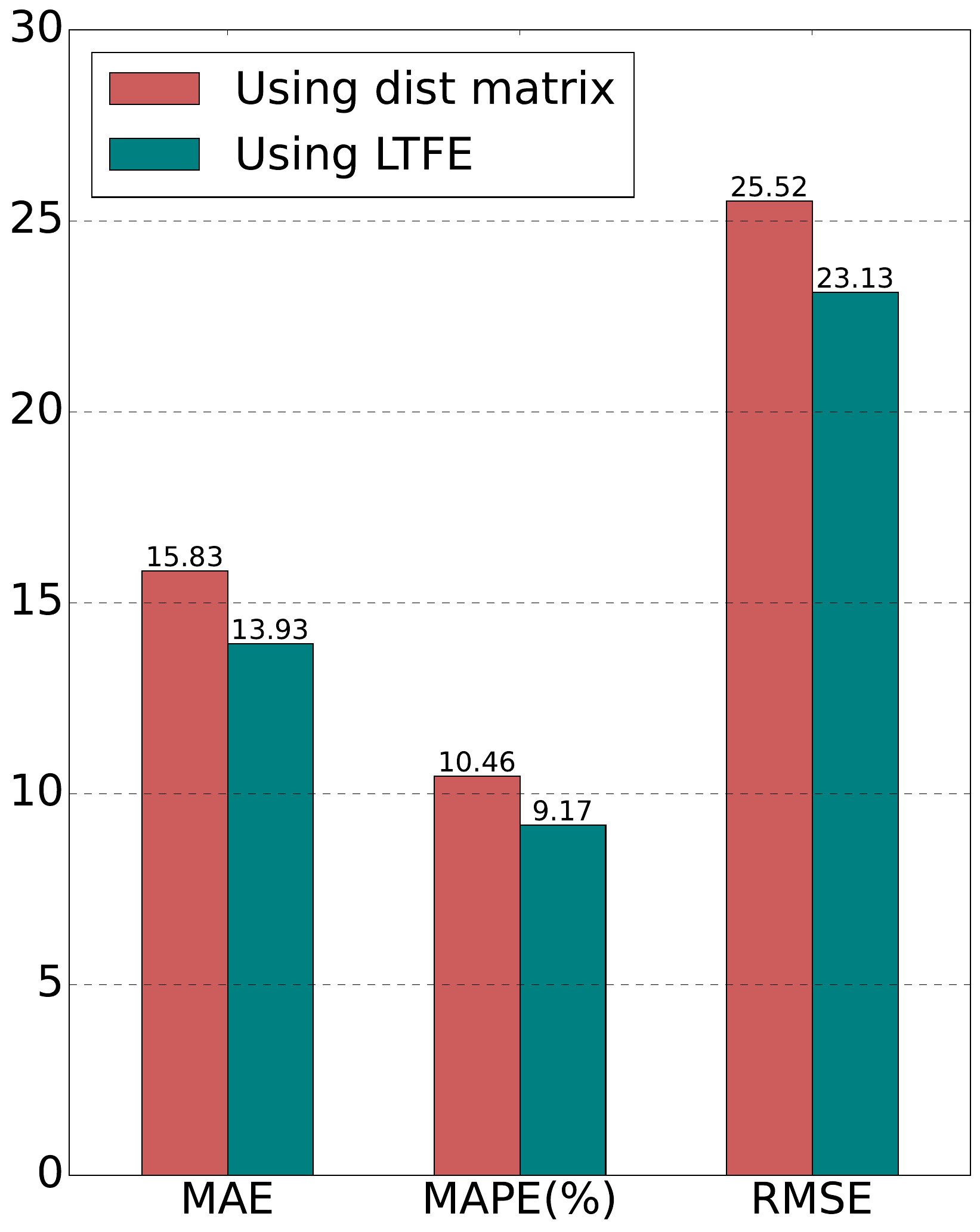}
    \caption{The comparison of graph structure replacement on PEMS03 (upper left), PEMS04 (upper right), PEMS07 (lower left), and PEMS08 (lower right).}
    \label{distance}
\end{figure} 

\begin{figure*}
    \centering
    \includegraphics[width=0.4952\columnwidth]{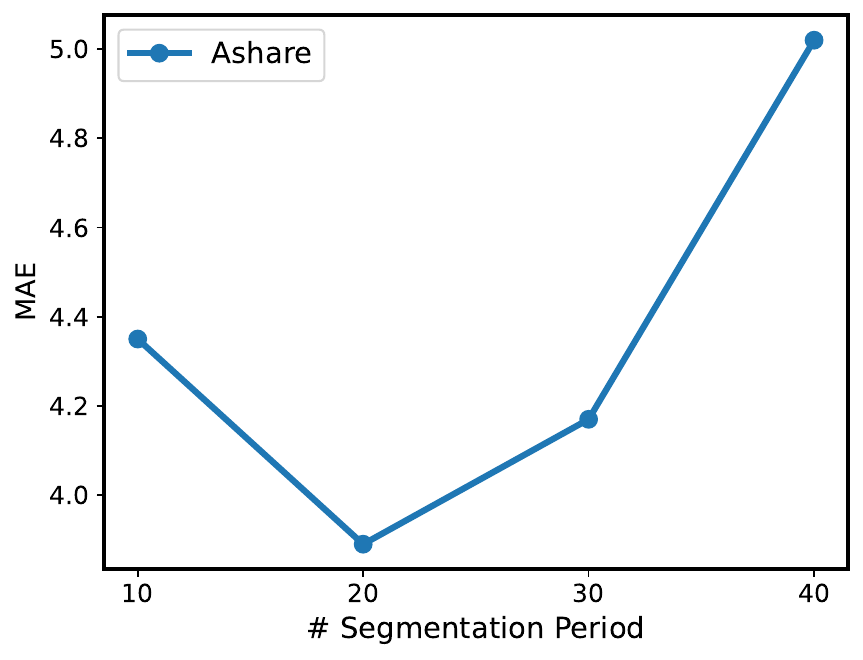}\includegraphics[width=0.4952\columnwidth]{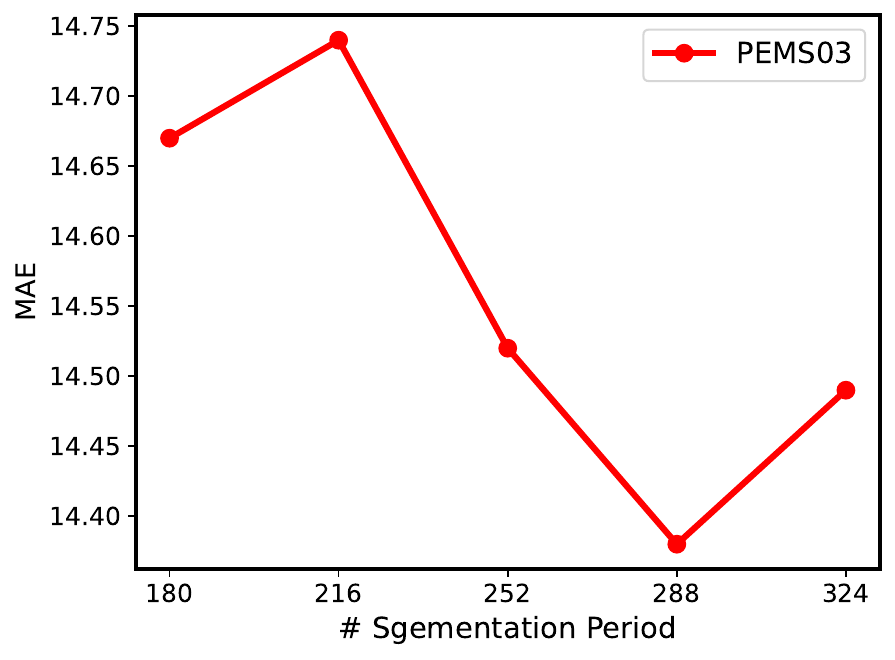}\includegraphics[width=0.4952\columnwidth]{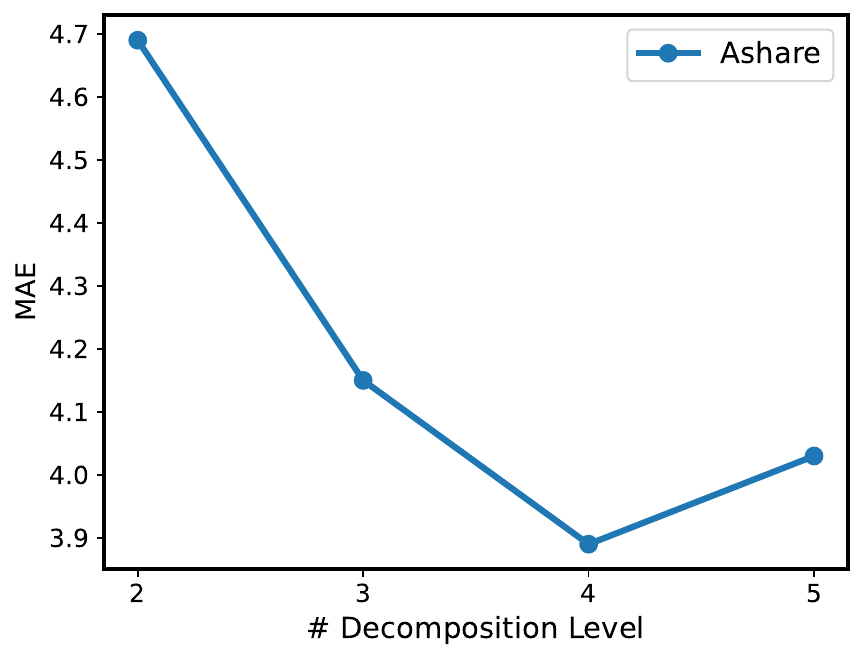}\includegraphics[width=0.4952\columnwidth]{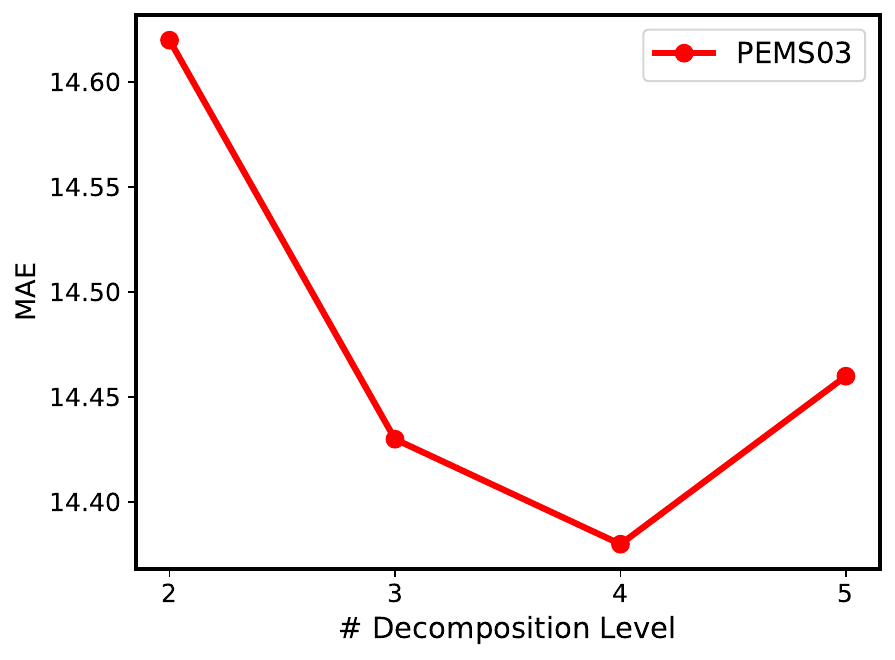}

    \includegraphics[width=0.4952\columnwidth]{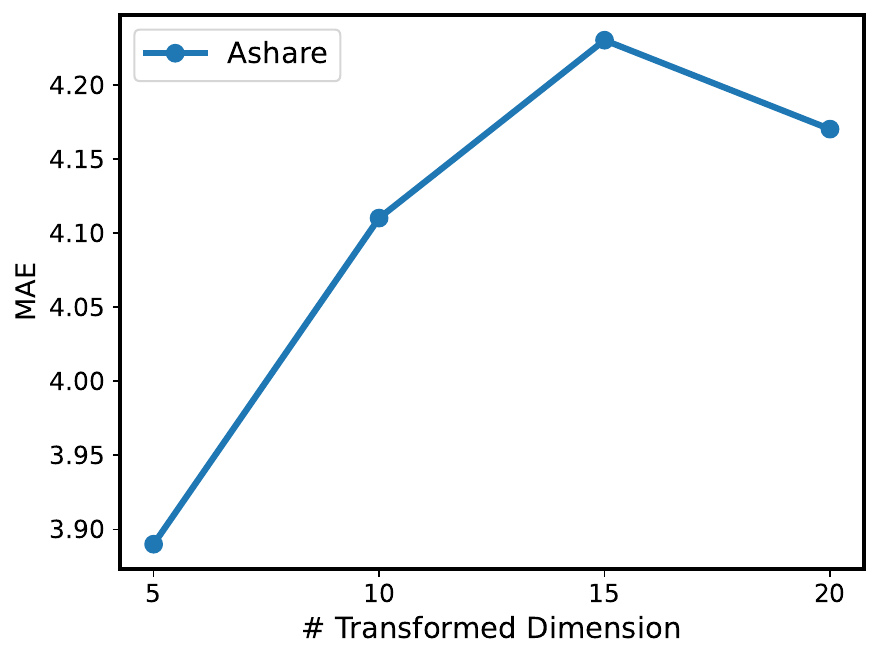}\includegraphics[width=0.4952\columnwidth]{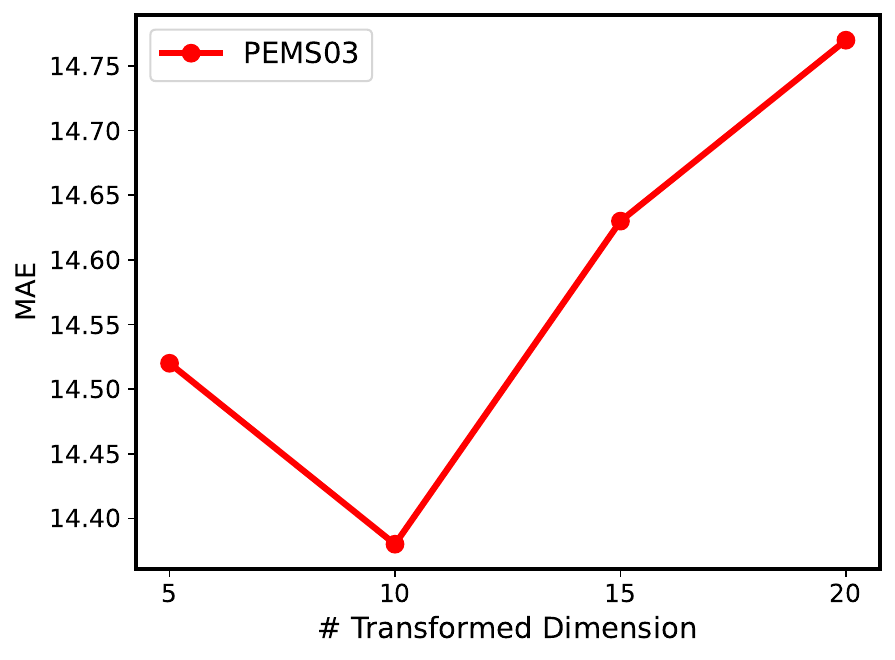}\includegraphics[width=0.4952\columnwidth]{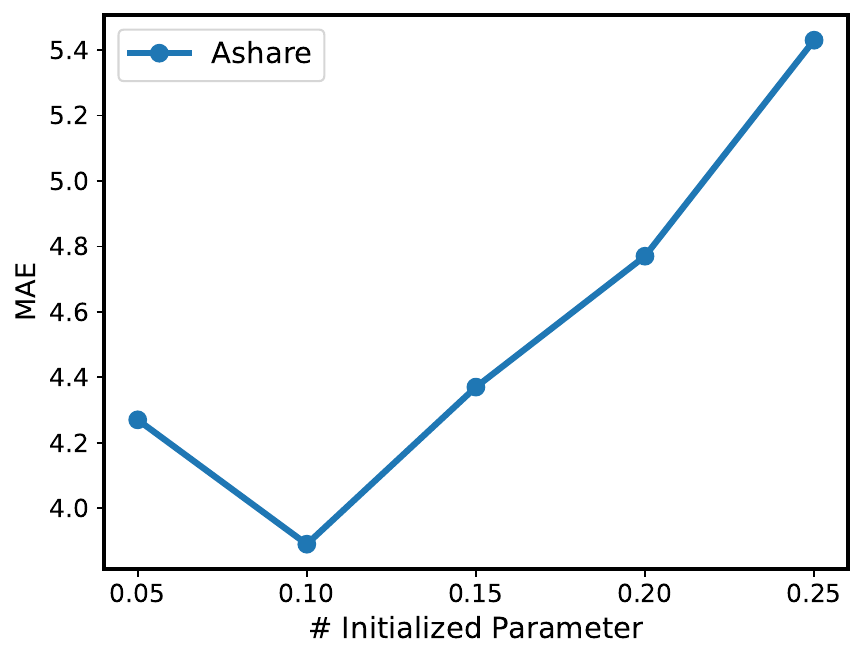}\includegraphics[width=0.4952\columnwidth]{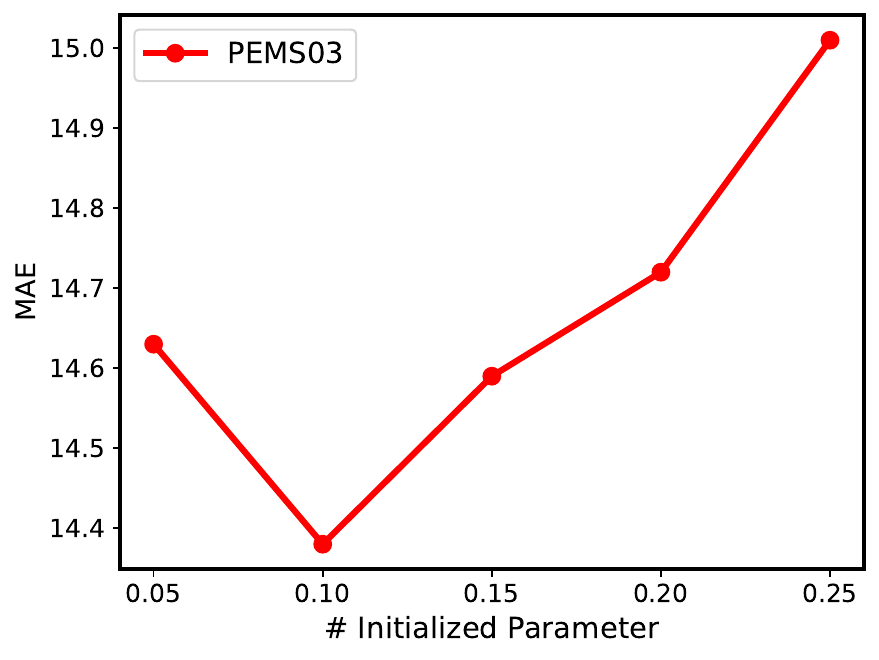}
    \caption{The parameter sensitivity experiments of the proposed FCDNet with respect to MAE on two
    datasets (Ashare and PEMS03).}
    \label{HyperStudy}
\end{figure*} 
\subsection{Hyperparameter Study}
Here we study the hyperparameters of FCDNet. Specifically, we focus on the number of decomposition level $L$ and segmentation period $P$ in LTFE, the number of transformed dimension $F$ in STFE, and
the initialized value of $\eta$. The experimental results on two datasets (PEMS03 and Ashare) are shown in Fig. 5. Note that when studying the effect of one hyperparameter, others are kept as the default values. Generally, the proposed FCDNet is not sensitive to
changes in hyperparameters. As can be seen from the data, changing the number of $P$ and the
number of $F$ does not influence the performance very much, and in most cases (except the cases when
we use large $\eta$ like 0.25), the
change in MAE is minimal. This
shows the robustness of our model. We vary the number of decomposition levels in the range of $\left\{2, 3, 4, 5\right\}$. Finding the right balance for the parameter $L$ is essential. If $L$ is set too small, it becomes challenging to distinguish and capture effective low-frequency signals essential for learning long-term stable correlations. Conversely, if $L$ is excessively large, valuable signals are lost, leading to a reduction in the learned long-range relational semantic information. Therefore, selecting an appropriate value for $L$ is a critical decision to ensure that the model effectively captures the relevant information without sacrificing signal quality.
Furthermore, it's worth noting that the substantial variance in hyperparameter settings (e.g., $F$) between the Ashare and PEMS03 datasets can be attributed to their distinct natures. The Ashare dataset belongs to the financial domain, while the PEMS03 dataset is associated with transportation. These fundamental differences in data domains require tailored hyperparameter configurations to accommodate the unique characteristics and challenges posed by each dataset.

\subsection{Dependency Graph Analysis}
\begin{figure}[H]
    \centering
    \vspace{-0.3cm}
    
    \includegraphics[width=0.495\columnwidth]{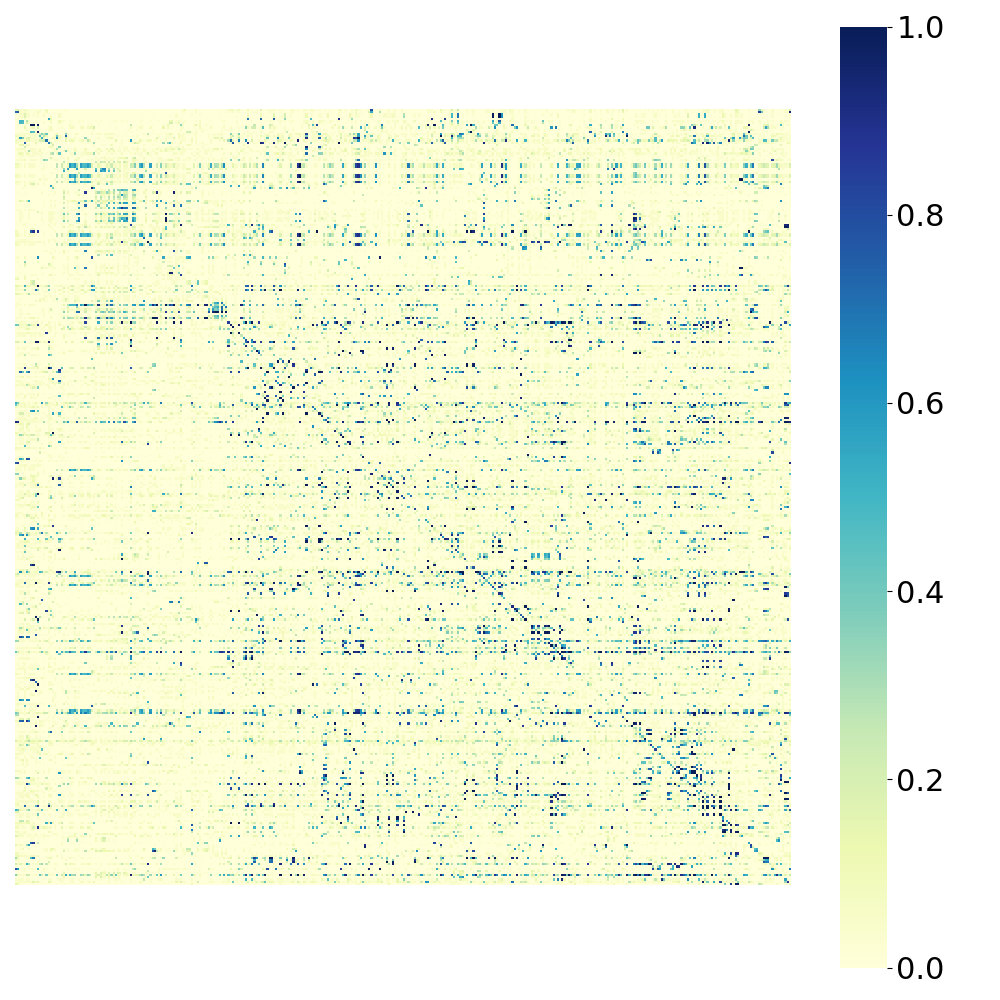}
    \includegraphics[width=0.495\columnwidth]{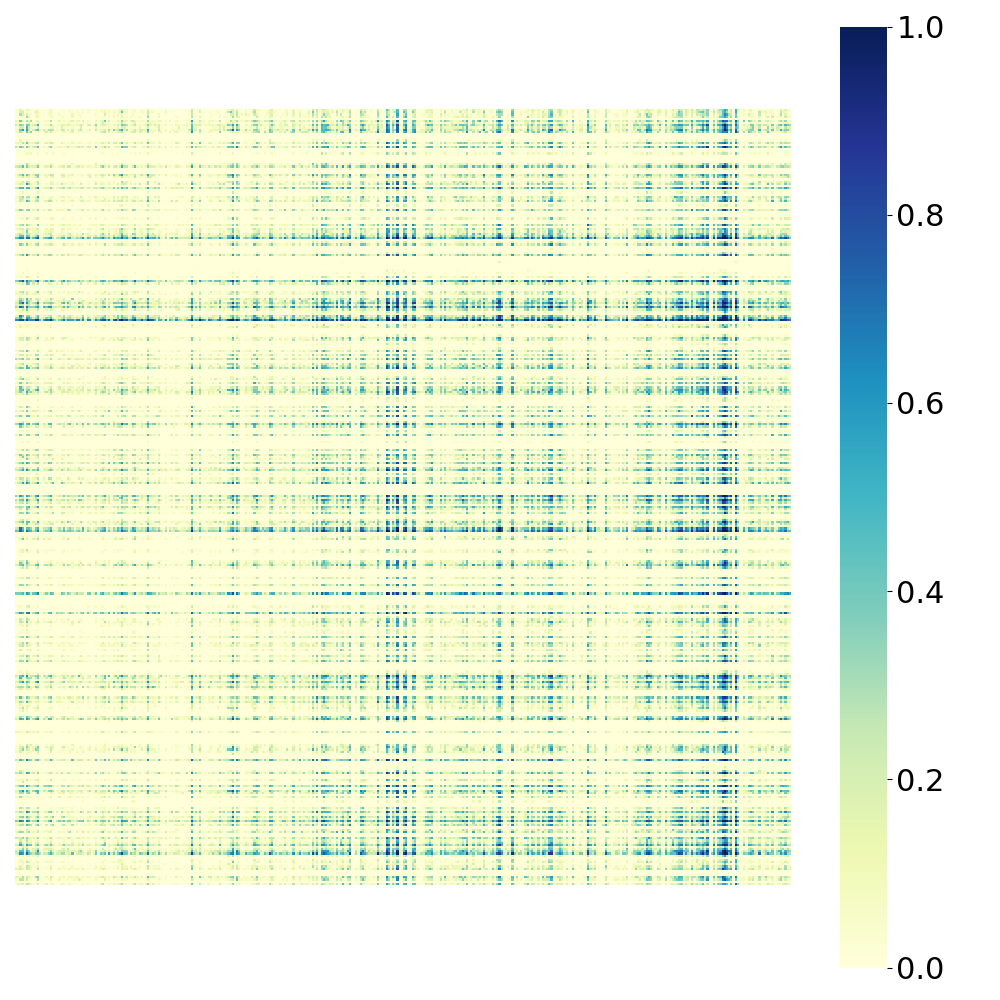}
    \caption{The final states of $A_{LF}$ (left), $A_{HF}$ (right) on PEMS03.}
    \label{heat}
\end{figure}

In this section, we look closer into the effectiveness of our proposed dependency modeling strategy. Figure~\ref{distance} illustrates the comparison of effects observed when replacing matrix $A_{LF}$ with the specific sensor distance matrix on the PEMS datasets. The figure clearly demonstrates that altering the learned $A_{LF}$ has a profound influence on the model's performance. This observation implies that matrix $A_{LF}$ encapsulates a substantial amount of semantic information, surpassing the content available in the distance matrix. The presence of this semantic information plays a pivotal and intuitive role in enhancing both the prediction accuracy and stability of the model. It underscores the significance of capturing and preserving semantic details for achieving accurate and consistent predictions in the given context. Figure~\ref{heat} visualizes the dependencies among different nodes on the PEMS03 dataset, where the left half represents $A_{LF}$ and the right half represents $A_{HF}$ obtained after model training.  It can be seen that the learned structure contains a variety of patterns. The diagonal of the heatmap for $A_{LF}$ reveals that the distance pattern is one of the most distinct and important features. This also means that the relatively invariant correlation between variables can be well discovered in the low-frequency information of long-term historical MTS. The heatmap for $A_{HF}$ pays more attention to the dynamic evolving correlation between variables, reflecting the shift of dynamic correlations in a certain period. The observation result is also similar to the phenomenon of evening correlation in traffic system~\cite{Poly}. Through analysis, it shows that the time-evolving correlation between variables can be effectively captured by the STFE module.
\section{Conclusion}
In this paper, we propose a novel model FCDNet that joins complementary dependency modeling and MTS forecasting. Unlike previous models that center on capturing static or dynamic relationships in multivariate time-series, our model focuses on capturing more complementary and essential long short-term static-dynamic dependencies between multivariate time-series from the frequency perspective. Specifically, FCDNet extracts the frequency information from long- and short-term MTS for constructing stable static and dynamic evolving dependency graphs. In addition, we migrate graph filters with stronger expressive power to the information fusion of structural information and time-series representation. Experiments on six MTS datasets show that FCDNet achieves state-of-the-art performance with fewer parameters compared to many strong baselines. The well-trained embeddings and learned dependencies can also be applied to other tasks. 






\bibliographystyle{elsarticle-num.bst}
\bibliography{NN}
\end{document}